\pdfoutput=1

\documentclass[11pt]{article}

\usepackage[preprint]{acl}

\usepackage{times}
\usepackage{latexsym}

\usepackage[T1]{fontenc}

\usepackage[utf8]{inputenc}

\usepackage{microtype}

\usepackage{inconsolata}

\usepackage{graphicx}
\usepackage{amsmath}
\usepackage{todonotes}
\usepackage{booktabs}
\usepackage{caption}
\usepackage{subcaption}
\usepackage[linesnumbered,ruled,vlined]{algorithm2e}
\SetKwRepeat{Do}{do}{while}

\title{Dictionaries to the Rescue: Cross-Lingual Vocabulary Transfer\\for Low-Resource Languages Using Bilingual Dictionaries}

\author{
  \textbf{Haruki Sakajo\textsuperscript{1}}, 
  \textbf{Yusuke Ide\textsuperscript{1}}, 
  \textbf{Justin Vasselli\textsuperscript{1}},\\
  \textbf{Yusuke Sakai\textsuperscript{1}}, 
  \textbf{Yingtao Tian\textsuperscript{2}}
  \textbf{Hidetaka Kamigaito\textsuperscript{1}}, 
  \textbf{Taro Watanabe\textsuperscript{1}}
\\
  \textsuperscript{1}Nara Institute of Science and Technology (NAIST), \textsuperscript{2}Sakana AI
\\
\texttt{sakajo.haruki.sd9@naist.ac.jp}\\
\texttt{\{sakai.yusuke.sr9, kamigaito.h, taro\}@is.naist.jp}
}

\begin{document}
\maketitle
\begin{abstract}
Cross-lingual vocabulary transfer plays a promising role in adapting pre-trained language models to new languages, including low-resource languages.
Existing approaches that utilize monolingual or parallel corpora face challenges when applied to languages with limited resources.
In this work, we propose a simple yet effective vocabulary transfer method that utilizes bilingual dictionaries, which are available for many languages, thanks to descriptive linguists.
Our proposed method leverages a property of BPE tokenizers where removing a subword from the vocabulary causes a fallback to shorter subwords.
The embeddings of target subwords are estimated iteratively by progressively removing them from the tokenizer.
The experimental results show that our approach outperforms existing methods for low-resource languages, demonstrating the effectiveness of a dictionary-based approach for cross-lingual vocabulary transfer\footnote{Code is available at \url{https://github.com/sj-h4/dict_trans_tokenizer}.}.
\end{abstract}

\section{Introduction}

Vocabulary transfer, or adaptation, is a method to bridge the gap between languages in language models~\cite{wang-etal-2020-extending, chau-etal-2020-parsing} to improve their performance for new languages.
Vocabulary transfer can also address token over-fragmentation~\cite{ahia-etal-2023-languages, petrov2023language, yamaguchi-etal-2024-empirical, yamaguchi2024effectivelyexpandvocabularyllms, han2025adapters}, where a sentence is split into spuriously many tokens, causing slower inference and more significant cost.
Previous works rely on monolingual or parallel corpora to align source and target languages or focus on vocabulary overlaps between them~\cite{minixhofer-etal-2022-wechsel, dobler-de-melo-2023-focus, pham-etal-2024-unibridge, liu-etal-2024-ofa,remy-delobelle2024transtokenization, minixhofer-etal-2024-zeroshot, han2025adapters, moroni-etal-2025-optimizing}.
However, these methods face a significant challenge when dealing with low-resource languages, where parallel corpora are often unavailable~\cite{artetxe-etal-2017-learning, fang-cohn-2017-model}, or when there is little lexical overlap between source and target languages that use different scripts.

\begin{figure}[t]
    \centering
    \includegraphics[width=0.94\linewidth]{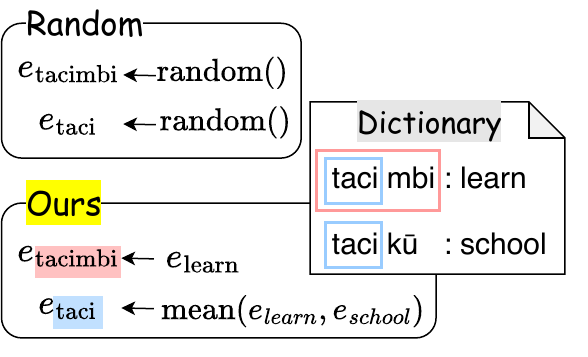}
    \caption{Conceptual illustration of our proposed method. Embeddings of target subwords are initialized using those of corresponding source subwords.}
    \label{fig:overview}
\end{figure}

Dictionaries offer a potential solution to this issue. %
When reading in an unfamiliar language, bilingual dictionaries can aid in comprehension by connecting words in the unknown language to their equivalents in a known language, and they can also enhance the performance of language models~\cite{vasselli-etal-2025-leveraging}.
Such dictionaries are often available, even when other linguistic resources are scarce, thanks to descriptive linguists, who produce them as part of their work in documenting languages~\cite{adams-etal-2017-cross}.

In this light, we propose an embedding initialization method for cross-lingual vocabulary transfer based on bilingual dictionaries as illustrated in Figure~\ref{fig:overview}.
Our proposed method utilizes dictionaries to map as many target subwords as possible to source subwords for embedding initialization.
This method is inspired by Trans-Tokenizer~\cite{remy-delobelle2024transtokenization} that maps target subwords to source subwords through word alignment using parallel corpora.
Our proposed method differs from Trans-Tokenizer in that we utilize dictionaries, which are more available for low-resource languages and have helpful features for subword mapping, and we estimate the target embeddings iteratively.
To perform iterative mapping, we utilize the BPE tokenizer algorithm~\cite{sennrich-etal-2016-neural}, but progressively remove subwords, forcing the tokenizer to decompose words into shorter and shorter subwords.
We estimate the embeddings for as many target subwords as possible by removing subwords mapped to source subwords from the vocabulary and reapplying tokenization.
The experimental results demonstrate that our dictionary-based approach outperforms existing methods and multilingual language models in terms of F1 score on a downstream task and perplexity, across low-resource languages with different scripts.
Our finding is that our dictionary-based method can improve the performance in low-resource languages that are genealogically distant from English and show typologically isolating or agglutinative characteristics (e.g., Uyghur, Khmer, and Manchu).

\section{Background and Related Work}
Cross-lingual vocabulary transfer consists of subword embedding initialization and language-adaptive pre-training.

\paragraph{Embedding Initialization.}
Embedding initialization is a fundamental component of cross-lingual vocabulary transfer.
\textsc{Wechsel}~\cite{minixhofer-etal-2022-wechsel} initializes target language subword embeddings using static embeddings obtained from monolingual corpora and bilingual dictionaries between source and target languages.
In contrast, \textsc{Focus}~\cite{dobler-de-melo-2023-focus} initializes target subword embeddings without bilingual dictionaries by leveraging subword overlap between source and target languages and static subword embeddings.
UniBridge~\cite{pham-etal-2024-unibridge} explores vocabulary size optimization and initializes subword embeddings, considering both syntactic and semantic aspects.
While it utilizes subword overlap for initialization, it also employs static embeddings from monolingual corpora to consider similarity for languages with rare subword overlap.
Trans-Tokenizer~\cite{remy-delobelle2024transtokenization} initializes target subword embeddings using word-level alignment from parallel corpora.
ZeTT~\cite{minixhofer-etal-2024-zeroshot} trains a hypernetwork that predicts the embeddings for target tokenizers with zero-shot.
However, each approach has limitations for low-resource languages: \textsc{Wechsel} requires substantial monolingual corpora and bilingual dictionaries, \textsc{Focus} and UniBridge face challenges with languages using different scripts or having minimal subword overlap with the source language, and Trans-Tokenizer is not applicable to languages lacking parallel corpora with the source language.

\paragraph{Language Adaptive Pre-Training (LAPT).}
After embedding initialization, most approaches train the embeddings or all weights in a target model.
This process is called Language-Adaptive Pretraining (LAPT)~\cite{chau-etal-2020-parsing} or continuous pre-training.
\textsc{Focus}, UniBridge, Trans-Tokenizer, and \citet{yamaguchi2024effectivelyexpandvocabularyllms} initialize subword embeddings with each approach, followed by LAPT.
UniBridge incorporates Masked Language Model (MLM) loss and KL divergence in its LAPT process, and others use MLM loss or Causal Language Model (CLM) loss.
Trans-Tokenizer and \citet{yamaguchi2024effectivelyexpandvocabularyllms} train the top and bottom two layers in CLMs with LoRA~\cite{hu2022lora}.
\citet{yamaguchi2024effectivelyexpandvocabularyllms} also demonstrates that continuous pre-training with multi-token prediction~\cite{pmlr-v235-gloeckle24a} sometimes improves the performance and does not cause performance degradation.

\section{Proposed Method}

\begin{figure*}[t]
    \centering
    \includegraphics[width=\linewidth]{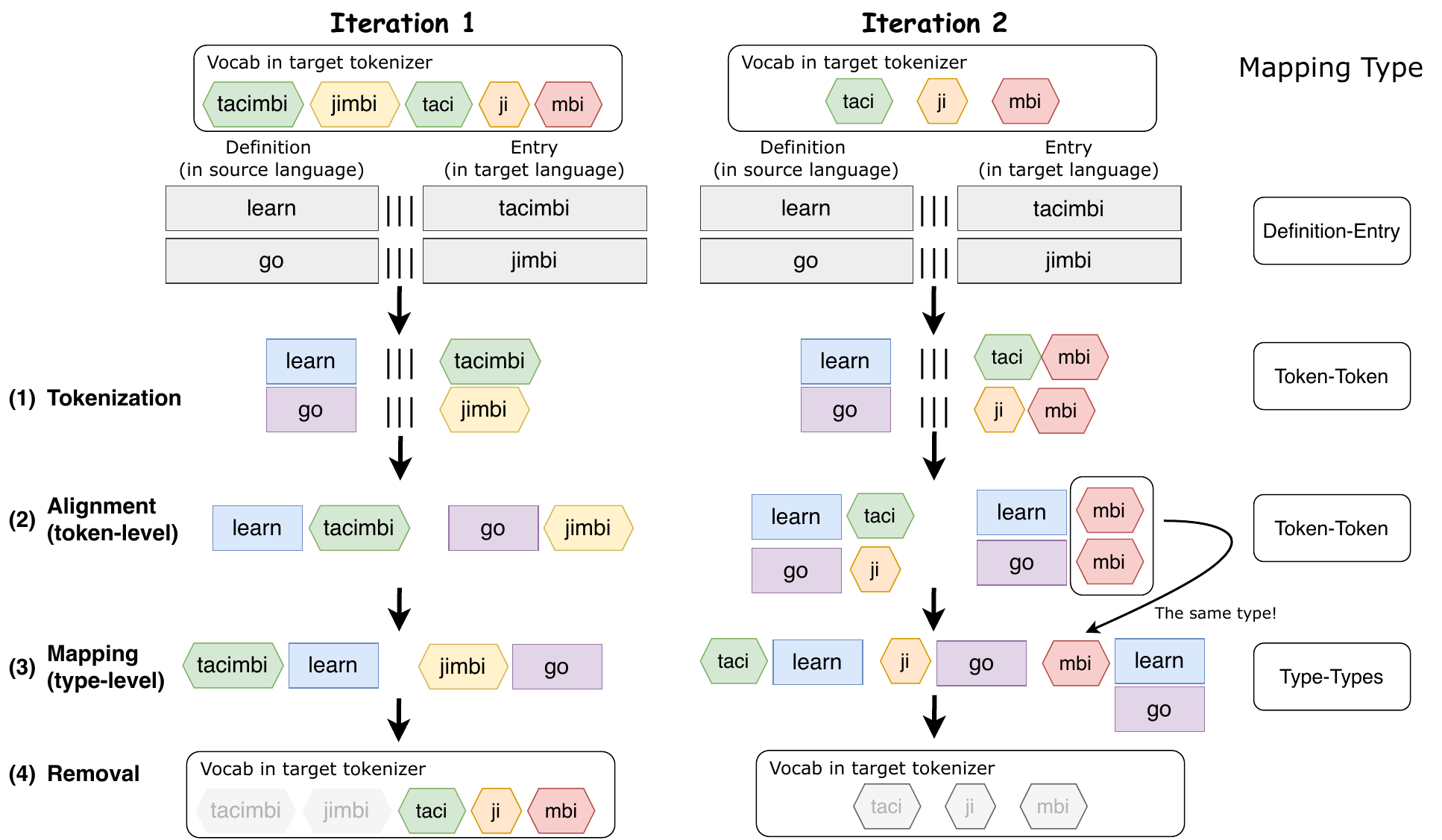}
    \caption{An illustration of the subword mapping part in the proposed method. We (1) tokenize definitions and entries, (2) align subwords using tokenized definition-entry pairs, (3) map a target subword to source subwords based on the alignment result, and (4) remove mapped target subwords from a target tokenizer. We repeat these processes until no target subwords are newly mapped. The target language is Manchu in this illustration, and ``-mbi'' is a vowel suffix.}
    \label{fig:method}
\end{figure*}

In this study, we propose a simple approach for low-resource languages utilizing bilingual dictionaries, which comprise of words (entries) in a target language and their definitions in a source language.
In this approach, we suppose that source models are trained with high-resource languages and have tokenizers with pre-trained embeddings.
The proposed method consists of three parts: training a tokenizer for the target language, aligning subwords between languages, and initializing target language subword embeddings.

\subsection{Tokenizer Training}
We train byte-level BPE tokenizers~\cite{sennrich-etal-2016-neural} using dictionary entries to obtain a tokenizer for the target language.
We employ byte-level BPE tokenizers to ensure that no tokens are out of vocabulary, even when training on limited resources.

\subsection{Subword Mapping}

The subword mapping algorithm is illustrated in Algorithm~\ref{alg:align_tokens}.
This process aims to maximize the number of target subwords that can be mapped to source subwords, reducing the number of unmapped subwords. Since unmapped target subwords are assigned the unknown token's embedding from the source language, our goal is to minimize their occurrence by increasing the coverage of mapped target subwords as much as possible.
This mapping process consists of four steps: tokenization, alignment, mapping, and removal, as illustrated in Figure~\ref{fig:method}.

\begin{algorithm}[!t]
\caption{Subword Mapping}
\label{alg:align_tokens}
\KwIn{
  \(S\): source tokenizer \\
  \(T\): target tokenizer \\
  \(\mathcal{D}\): dictionary (entry-definition pair) \\
  \(\mathcal{M}\): target-source subwords mappings (one-to-many) and the count for each mapping \\
}
\KwOut{Target-Source one-many subword mapping}
\BlankLine
\Do{\(|\mathcal{M}_{\text{new}}| > 0\)}{
\(\mathcal{C} \leftarrow \text{TokenizeCorpus}(S,\, T,\, \mathcal{D})\)\;
\(\mathcal{A} \leftarrow \text{FastAlign}(\mathcal{C})\)\;
\( \mathcal{M}_\text{new} \leftarrow \text{MapSubwords}(\mathcal{A})\)\;
\(\mathcal{M} \leftarrow \text{UpdateMapping}(\mathcal{M}, \, \mathcal{M}_{\text{new}})\)
\BlankLine
\(T \leftarrow \text{DeleteTokens}(T,\, \mathcal{M}_{\text{new}})\)\;
}
\Return \(\mathcal{M}\)\;
\end{algorithm}

\paragraph{(1) Tokenization.}
First, we tokenize dictionary entries and definitions using the trained target language tokenizer and the source model's tokenizer, respectively (line 2 in Algorithm~\ref{alg:align_tokens}).

\paragraph{(2) Alignment.}
We treat the entry-definition pairs as a parallel corpus and employ fast\_align~\cite{dyer-etal-2013-simple}, which is based on the IBM translation models~\cite{brown-etal-1993-mathematics}, IBM Model 2, as an alignment tool to obtain target-source subword pairs and their counts (line 3).
In this step, the subwords are handled as tokens, not types.

\paragraph{(3) Mapping.}
Based on the target-source subword pairs obtained in step 2, we create a one-to-many mapping for all target language subwords at the type level (line 4), and then update the subword mapping (line 5).
The update process includes updating target-source subword pairs at the type level and aggregating their count.

\paragraph{(4) Removal.}
After creating target-source subword mappings, we remove the mapped target subwords from the target tokenizer's vocabulary and the merged pairs that contain the subwords (line 6).
The aim is to assign source subwords to unmapped target subwords in the next loop.
Since BPE tokenizers iteratively merge subwords into longer subwords, only the longest subwords are mapped by default, and their constituent subwords remain unmapped if this removal is skipped.
Removing subwords enables the mapping of shorter subwords, as BPE tokenizers fall back to shorter segments when a subword is no longer available in the vocabulary.

We repeat these steps until no new subwords are mapped.
We assign the UNK token from the source language model for target language subwords without corresponding source language subwords in the tokenizer's vocabulary.

This subword mapping method is different from the Trans-Tokenizer.
Our proposed method employs subword-level alignment because dictionaries are primarily composed of words, and the words are short.
It also estimates mappings for shorter subwords, while the Trans-Tokenizer employs word-level alignment and estimates mapping heuristically.

\subsection{Initializing Subword Embeddings}
Let \(A_t\) be the set of source language subwords corresponding to a target language subword  \(t\), \(E_s^S\) and \(E_t^T\) be the sets of embeddings for a source language subword \(s\) and a target language subword \(t\), respectively.
Let \(c(x, y)\) be the count of a pair of \(x\) and \(y\).
The relative counts of subword \(s\) in \(\mathcal{M}_t\) is defined as:
\begin{equation*}
c(s|t) = \frac{c(s, t)}{\Sigma{x \in \mathcal{M}_t}c(x, t)}
\end{equation*}
Using the embedding \( \boldsymbol{e}_s^S \in E^S\) of the source subword \(s \in A_t\), the embedding \( \boldsymbol{e}_t^T \in E^T\) of the target subword \(t\) is initialized as a weighted average of corresponding source subwords embeddings\footnote{Note that Llama 3 tokenizer does not have the UNK token, so we initialize the subword randomly.}:
\begin{equation*}
\boldsymbol{e}_t^T = \sum{s \in \mathcal{M}_t} c(s|t) \cdot \boldsymbol{e}_s^S
\end{equation*}
We finally copy the embeddings of the special tokens, digits, and punctuations from the source model to the target model and then replace the embeddings in the source model with the initialized embeddings.

This embedding initialization method is different from \textsc{Focus}.
\textsc{Focus} initializes target subword embeddings using source-target subword mapping obtained from overlapping subwords between source and target languages and static embeddings from FastText~\cite{bojanowski-etal-2017-enriching}.
In contrast, our method uses the mapping trained with dictionaries.

\section{Experimental Setups}

\paragraph{Languages.}
We use English as a source language and German, Japanese, Old English, Uyghur, Sanskrit, Khmer, and Manchu as target languages.
We treat German and Japanese as high-resource languages and other languages as low-resource languages.
As source models, we use RoBERTa~(base)~\cite{liu2019robertarobustlyoptimizedbert}, XLM-R~(base)~\cite{conneau-etal-2020-unsupervised}, Llama 3.1 8B~\cite{grattafiori2024llama3herdmodels}, and Gemma 2 9B~\cite{gemmateam2024gemma2improvingopen}.
The detailed information about the models is provided in Appendix~\ref{sec:detai_experimental_setups}

\begin{table}[t]
    \centering
    \begin{tabular}{llr}
    \toprule
        Language & Script & Entries \\
    \midrule
        German & Latin & 101,997 \\
        Japanese & Kanji, Kana & 25,969 \\
        Old English & Latin & 7,592 \\
        Uyghur & Arabic & 1,131 \\
        Sanskrit & Devanagari & 5,282 \\
        Khmer & Khmer & 5,656 \\
        Manchu & Latin & 21,620 \\
    \bottomrule
    \end{tabular}
    \caption{The script and the number of entries in the dictionary for each language.}
    \label{tab:dataset_size}
\end{table}

\begin{table}[t]
    \centering
    \resizebox{\linewidth}{!}{
    \begin{tabular}{lrrrr}
    \toprule
         & \multicolumn{3}{c}{\textsc{Focus}} & Ours\\
        \cmidrule(lr){2-4}
        Language & XLM &  Llama 3.1 &  Gemma 2 &  \\
    \midrule
        German & 1,948,497 & 2,063,441 & 1,720,272 & 2,282,288 \\
        Japanese & 2,320,596 & 2,554,873 & 2,275,944 & 4,291,810 \\
        Old English & 398,508 & 416,885 & 327,382 & 660,516 \\
        Uyghur & 1,440,550 & 1,582,594 & 1,359,842 & 3,656,931 \\
        Sanskrit & 1,373,894 & 2,851,798 & 1,142,922 & 4,612,928 \\
        Khmer & 1,983,542 & 4,203,481 & 1,955,756 & 7,589,786 \\
        Manchu & 191,649 & 180,748 & 77,737 & 209,420 \\
    \bottomrule
    \end{tabular}
    }
    \caption{The number of tokens in the data for LAPT for each language.}
    \label{tab:dataset_corpus_size}
\end{table}

\begin{table*}[t]
    \centering
    \begin{tabular}{lrrrrrrr}
    \toprule
         & German & Japanese & Old English & Uyghur & Sanskrit & Khmer & Manchu \\
    \midrule
        RoBERTa & \underline{89.61} & \underline{75.33} & \textbf{62.39} & 38.73 & \underline{51.48} & 27.58 & 73.52 \\
        XLM-R & \textbf{90.27} & \textbf{81.28} & 37.59 & 28.30 & 48.85 & 34.78 & 65.32 \\
        \hline
        \textsc{Focus} (XLM-R) & 89.28 & 77.22 & 37.57& 23.00  & 36.16 & 19.35  & 28.00 \\
        \hspace{1em}+ LAPT & 90.00 & 77.46 & 37.57  & 37.16 & 12.33 & 12.33 & 28.03 \\
        \hline
        Ours (RoBERTa) & 76.99  & 71.93 & 45.43 & 36.06 & 44.23 & 42.21 & \textbf{94.87} \\
        \hspace{1em}+ LAPT & 76.43 & 73.60 & \underline{52.71} &  \textbf{64.52}  & 42.08 & \textbf{62.96} & 92.87 \\
        Ours (XLM-R) &  75.98 & 72.66 & 35.10 & 25.07 & 37.53 & 36.55 & \underline{94.69} \\
        \hspace{1em}+ LAPT & 75.98 & 74.73 & 40.94  & \underline{59.41} & \textbf{56.99} & \underline{58.37} & 91.39 \\
    \bottomrule
    \end{tabular}
    \caption{Micro F1 scores of MLMs for WikiANN and ManNER.}
    \label{tab:result_ner}
\end{table*}

\paragraph{Evaluation Metrics.}
We evaluate the performance of our initialization method using F1 for the NER performance of a masked language model (RoBERTa) and perplexity for causal language models (Llama 3.1 and Gemma 2). 
The perplexity is normalized by word length, not subword length, for fair comparison across tokenizers.
This is because comparing models with different tokenizers based on the perplexity normalized by subword length is unfair~\cite{roh2020unigramnormalizedperplexitylanguagemodel}.
We compute the word length using MeCab~\cite{kudo-etal-2004-applying} with unidic-lite\footnote{\url{https://github.com/polm/unidic-lite}} for normalization for Japanese and tokenize by space for other languages.
We use this perplexity for comparison between models within the same language, but not for comparison across different languages.

\paragraph{Datasets.}
As dictionary data, we use different resources for each language.
For Manchu, we use data extracted from the Manchu–English dictionary~\cite{Norman2013-tq}\footnote{\url{https://github.com/tyotakuki/manchuvocabdata}}.
For German and Japanese, we employ the \text{<target language>}-English bilingual dictionary from MUSE~\cite{lample2017unsupervised} that is utilized in \textsc{Wechsel}~\cite{minixhofer-etal-2022-wechsel}\footnote{\url{https://github.com/facebookresearch/MUSE}}. 
For Uyghur, Sanskrit, and Khmer, we use the \text{<target language>}-English dictionary from Wiktionary as employed in \textsc{Wechsel}~\cite{minixhofer-etal-2022-wechsel}\footnote{\url{https://github.com/CPJKU/wechsel/tree/main/dicts}}.
Note that the Manchu dictionary includes entries and definitions that consist of multiple words, whereas in the other dictionaries, each target language word is paired with a single word.

We use ``Manwen Laodang''\footnote{The Manwen Laodang data used in this study  was collected in-house.} as a text corpus for initializing with \textsc{Focus} and training for LAPT for Manchu and Wikipedia dataset~\cite{wikidump}\footnote{\url{https://huggingface.co/datasets/wikimedia/wikipedia}. We use the 20231101 dump.} for other languages.
For NER, we use ManNER~\cite{lee-etal-2024-manner} for Manchu and WikiANN~\cite{pan-etal-2017-cross, rahimi-etal-2019-massively} for other languages.
Table~\ref{tab:dataset_size} shows the writing system and the number of entries in the dictionary for each language\footnote{Note that Manchu uses the Manchu alphabet, but the dictionary and dataset transcribe it in the Latin alphabet.}.

\begin{table*}[t]
    \centering
    \resizebox{\textwidth}{!}{
    \begin{tabular}{lrrrrrrr}
    \toprule
         & German & Japanese & Old English & Uyghur & Sanskrit & Khmer & Manchu \\
    \midrule
        Llama 3.1 & \underline{655.26} & 7868.14 & 199411567.86 & $2.07 \times 10^{24}$ & 548192867.74 & $\inf$ & $2.30 \times 10^{19}$  \\
        \textsc{Focus} & 13275123.06 & 58188456.24 & $1.03 \times 10^{10}$ & $4.36 \times 10^{22}$ & $2.99 \times 10^{20}$ & $\inf$ & 1893791141.42 \\
        \hspace{1em}+ LAPT & 2375.69 & 574.71 & 57810645.31 & $7.53 \times 10^{20}$ & 12376253.13 & $\inf$ & 1288173.19 \\
        Ours & 444876.62 & \underline{473.72} & \underline{46180.42} & \underline{18053.31} & \underline{1503.39} & \underline{64508.22} & \underline{144818.36} \\
        \hspace{1em}+ LAPT & \textbf{88.61} & \textbf{4.42} & \textbf{406.53} & \textbf{168.43} & \textbf{3.56} & \textbf{4.32} & \textbf{502.02} \\
        \midrule
        Gemma2 & \underline{2013.66} & 494072.35 & $3.36 \times 10^{11}$ & $3.98 \times 10^{27}$ & $1.02 \times 10^{10}$ & $\inf$ & $1.06 \times 10^{16}$ \\
        \textsc{Focus} & $2.83 \times 10^{11}$ & 993607.28 & $5.17 \times 10^{13}$ & $1.71 \times 10^{30}$ & $5.09 \times 10^{14}$ & $\inf$ & 1025102.29 \\
        \hspace{1em}+ LAPT & 14045.20 & \underline{356.75} & 57810645.31 & $2.89 \times 10^{23}$ & 144650900.73 & $\inf$ & \underline{42687.69} \\
        Ours & 6802328.30 & 1340.02 & \underline{4685467.63} & \underline{3163709.28} & \underline{97101.92} & \underline{652351.91} & 4848206.72 \\
        \hspace{1em}+ LAPT & \textbf{595.14} & \textbf{5.83} & \textbf{1324.49} & \textbf{83.32} & \textbf{10.57} & \textbf{16.80} & \textbf{509.64} \\
    \bottomrule
    \end{tabular}
    }
    \caption{Perplexity of CLMs. The perplexity is normalized by word length, not subword length, for fair comparison across different tokenizers.}
    \label{tab:result_ppl}
\end{table*}

\paragraph{Training.}

After embedding initialization, we conduct continuous pretraining with up to 3,000 samples for LAPT.
The number of tokens for each model is shown in Table~\ref{tab:dataset_corpus_size}.
For masked language models, we fine-tune all layers using the MLM loss.
We also fine-tune both models with LAPT and models without LAPT for the downstream task.
We train only the top and bottom two layers with LoRA~\cite{hu2022lora} in the CLMs, following \citet{remy-delobelle2024transtokenization} and \citet{yamaguchi2024effectivelyexpandvocabularyllms}.
We also adopt multi-token prediction~\cite{pmlr-v235-gloeckle24a} for training efficiency, which causes no significant performance degradation~\cite {yamaguchi2024effectivelyexpandvocabularyllms}. 
The hyperparameters are in Appendix~\ref{sec:detai_experimental_setups}.

\paragraph{Baselines.}

We use \textsc{Focus} as a baseline for embedding initialization performance because this is a widely used baseline in vocabulary transfer studies~\cite{remy-delobelle2024transtokenization, yamaguchi2024effectivelyexpandvocabularyllms, pham-etal-2024-unibridge, minixhofer-etal-2024-zeroshot}.
We train target tokenizers with up to 50,000 samples in the dataset.
We initialize target subword embeddings with \textsc{Focus}.
We use up to 50,000 samples in a dataset for initialization.
We apply \textsc{Focus} to XLM-R instead of RoBERTa as an MLM baseline because \textsc{Focus} performs better for multilingual models than monolingual models.
We also apply \textsc{Focus} to Llama 3.1 and Gemma 2 as CLM baselines.

\begin{figure}[t]
    \centering
    \begin{subfigure}{0.45\textwidth}
        \centering
        \includegraphics[width=\textwidth]{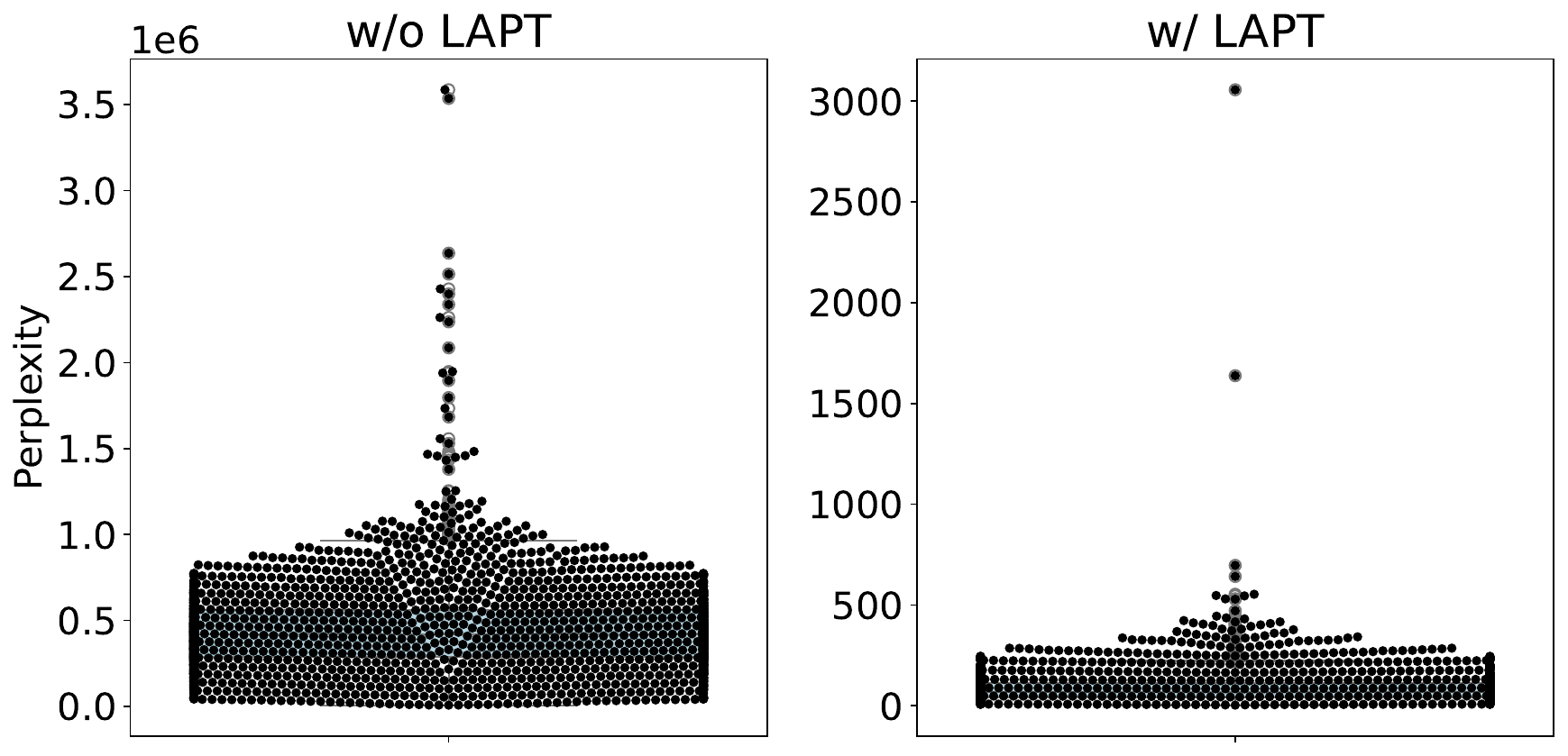}
        \caption{German}
    \end{subfigure}
    \hfill
    \begin{subfigure}{0.45\textwidth}
        \centering
        \includegraphics[width=\textwidth]{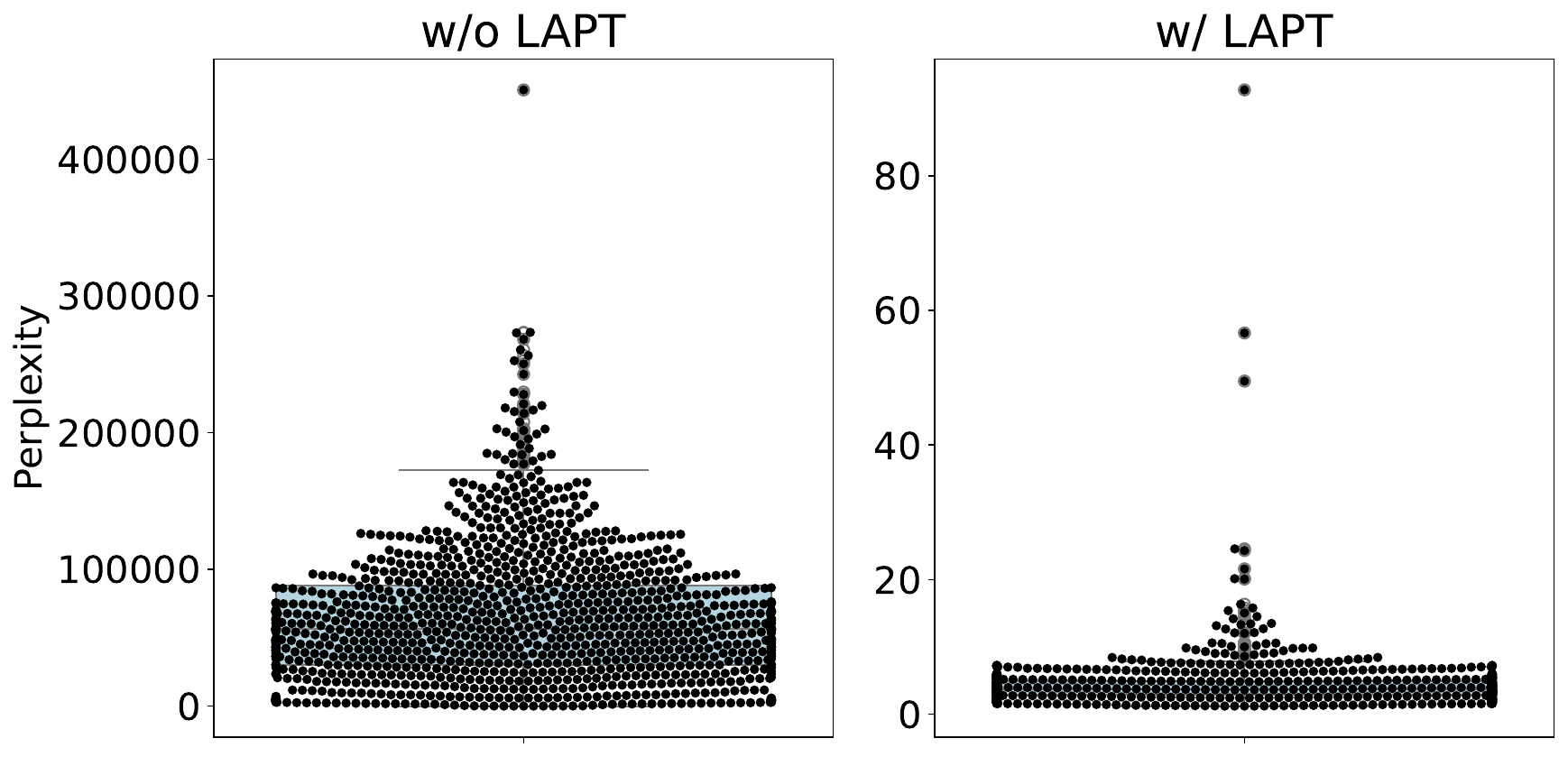}
        \caption{Khmer}
    \end{subfigure}
    \hfill
    \begin{subfigure}{0.45\textwidth}
        \centering
        \includegraphics[width=\textwidth]{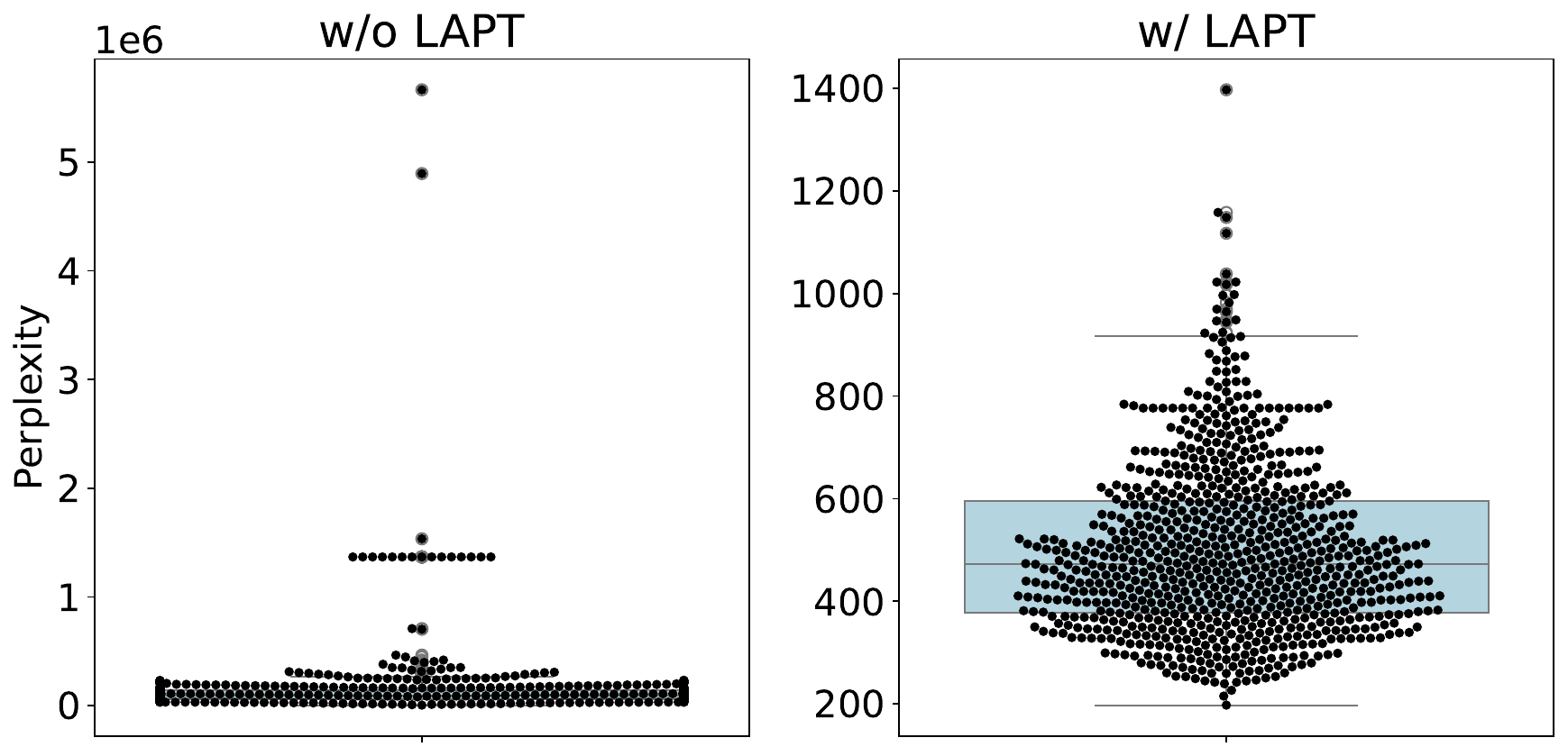}
        \caption{Manchu}
    \end{subfigure}
    \hfill
    \caption{The perplexity distribution and box plot of Llama 3.1-based our proposed method.}
    \label{fig:ppl_distribution}
\end{figure}

\section{Results and Discussion}
\label{sec:result}

Table~\ref{tab:result_ner} shows the results on NER.
When the results of RoBERTa-based models were compared with those of the proposed method, the proposed method achieved better performance for Uyghur, Khmer, Sanskrit, and Manchu.

Table~\ref{tab:result_ppl} shows the perplexity for the base and transferred models of Llama 3.1 and Gemma 2. 
This result indicates that the performance of the proposed method with LAPT achieves the best performance.
Figure~\ref{fig:ppl_distribution} illustrates the perplexity distribution of the Llama 3.1-based model for German and Manchu using our proposed method.
The distributions of other languages are illustrated in Figure~\ref{fig:ppl_distribution_all} in Appendix~\ref{sec:appendix_ppl_distribution}.
There are some outliers regardless of whether LAPT is performed or not, suggesting out-of-distribution~(OOD) issues occur.
We also visualize the embeddings in Appendix~\ref{sec:appendix_visualization}.

\subsection{Data Size}

Table~\ref{tab:comparison_data_size} shows the number of words of target languages in the data used for embedding initialization.
The results shown in Tables~\ref{tab:result_ner},~\ref{tab:result_ppl}, and ~\ref{tab:comparison_data_size} demonstrate that our proposed method achieves better performance than \textsc{Focus} in the low-resource languages, with less than 10\% of the data, highlighting the efficiency of the proposed method.

On the other hand, \textsc{Focus} performs better than our proposed method for high-resource languages, German and Japanese.
This indicates that \textsc{Focus} can effectively initialize subword embeddings if a sufficient amount of data is available, while our proposed method can initialize subword embeddings with a limited amount of data.

\begin{table}[t]
    \centering
    \begin{tabular}{lrr}
    \toprule
        Language & \textsc{Focus} & Ours \\
    \midrule
        German & 21,582,818 & 101,997 \\
        Japanese & 3,118,885 & 25,969 \\
        Old English & 353,515 & 7,592 \\
        Uyghur & 2,771,058 & 1,131 \\
        Sanskrit & 2,812,121 & 5,282 \\
        Khmer & 1,937,229 & 5,656 \\
        Manchu & 397,659 & 21,620 \\
    \bottomrule
    \end{tabular}
    \caption{The number of words in the data of target languages for embedding initialization. We count strings separated by spaces as a word.}
    \label{tab:comparison_data_size}
\end{table}

\begin{table}[t]
    \centering
    \resizebox{\linewidth}{!}{
    \begin{tabular}{lrr}
    \toprule
        Language & Tokenizer size & Mapped tokens \\
    \midrule
        German & 50,265 & 43,656 (86.85 \%) \\
        Japanese & 33,449 & 32,043(95.80 \%) \\
        Old English & 11,427 & 10,059 (88.03 \%)\\
        Uyghur & 2,693 & 2,372 (88.08 \%)\\
        Sanskrit & 1,719 & 1,672 (97.27 \%)\\
        Khmer & 1,634 & 1,465 (89.67 \%)\\
        Manchu & 20,578 & 15,918 (77.35 \%)\\
    \bottomrule
    \end{tabular}
    }
    \caption{The vocabulary size in tokenizers and the number of mapped tokens during transferring Llama 3.1.}
    \label{tab:tokenizer_vocab}
\end{table}

\begin{table*}[t]
  \centering
    \resizebox{\linewidth}{!}{
    \begin{tabular}{lrrrrrrrr}
    \toprule
         & RoBERTa & XLM-R & Llama 3.1 & Gemma 2 & \multicolumn{3}{c}{\textsc{Focus}} & Ours\\
        \cmidrule(lr){6-8}
        Language &  &  &  &  & XLM &  Llama 3.1 &  Gemma 2 &  \\
    \midrule
        German & 3,686,547 & 2,426,841 & 2,983,799 & 2,598,287 & 2,069,545 &  2,198,015 & 1,827,442 & 2,429,275 \\
        Japanese & 7,720,302 & 3,616,523 & 4,129,765 & 3,608,275 & 2,479,331 &  2,735,561 & 2,431,308 & 4,585,446 \\
        Old English & 100,320 & 83,383 & 93,820 & 86,062 & 46,263 & 47,782 & 37,764 & 75,784 \\
        Uyghur & 3,482,499 & 837,816 & 2,189,895 & 1,750,318 & 472,816 &  525,931 & 455,439 & 1,264,060 \\
        Sanskrit & 4,482,539 & 946,657 & 1,541,635 & 1,343,833 & 592,570 &  1,230,733 & 497,080 & 19,974,128 \\
        Khmer & 9,987,789 & 1,318,386 & 5,773,353 & 3,533,035 & 736,043 & 1,674,177 & 726,099 & 3,100,085 \\
        Manchu & 104,660 & 87,861 & 95,225 & 85,902 & 52,104 &  49,198 & 21,111 & 56,912 \\
    \bottomrule
    \end{tabular}
    }
    \caption{The number of subwords in the test data for perplexity for each language. Note that the same tokenizer, trained on dictionary entries, was used for each language for our proposed method during experiments.}
    \label{tab:num_subwords}
\end{table*}

\begin{table*}
    \centering
    \resizebox{\textwidth}{!}{
    \begin{tabular}{lrrrrrrrrrrrr}
    \toprule
        Base Model & \multicolumn{4}{c}{RoBERTa} & \multicolumn{4}{c}{Llama 3.1} & \multicolumn{4}{c}{Gemma 2}\\
        \cmidrule(lr){2-5} \cmidrule(lr){6-9} \cmidrule(l){10-13}
        & iter 1 & iter2 & iter 3 & iter 4 & iter 1 & iter2 & iter 3 & iter 4 & iter 1 & iter2 & iter 3 & iter 4 \\
    \midrule
        German & 43,591 & 62 & 3 & 0 & 44,134 &  63 & 1 & 0 & 44,652 & 63 & 1 & 0 \\
        Japanese & 21,003 & 10,987 & 34 & 0 & 21,003 & 11,008 & 34 & 0 & 21,002 & 11,013 & 28 & 0 \\
        Old English & 7,520 & 2,503 & 17 & 10 & 7,523 & 2,508 & 17 & 10 & 7,522 & 2,508 & 17 & 10 \\
        Uyghur & 1,152 & 1,079 & 109 & 27 & 1,152 & 1,079 & 109 & 27 & 1,152 & 1,079 & 109 & 27  \\
        Sanskrit & 1,575 & 90 & 4 & 2 & 1,575 & 93 & 3 & 0 & 1,575 & 94 & 3 & 0 \\
        Khmer & 1,370 & 77 & 3  & 0 & 1,385 & 78 & 2 & 0 & 1,384 & 76 & 3 & 0 \\
        Manchu & 15,053 & 846 & 4 & 0 & 15,066 & 848 & 4 & 0 & 15,068 & 851 & 4 & 0 \\
    \bottomrule
    \end{tabular}
    }
    \caption{The number of mapped subwords in the iteration until the fourth iteration.}
    \label{tab:mapped_iteration}
\end{table*}

\subsection{Mapped Subwords Coverage}
\label{sec:performance_ang_coverage}
Table~\ref{tab:tokenizer_vocab} shows the trained tokenizers' vocabulary size and the number of mapped subwords that were transferred from Llama 3.1 to target languages.
The coverage of mapped tokens varies by language and can be categorized into two groups: Japanese and Sanskrit, as well as others.
Japanese and Sanskrit have relatively regular stem changes.
On the other hand, German has complex inflection, and Uyghur and Manchu have vowel harmony, which makes tokenization inefficient.
These linguistic features cause the coverage of mapped tokens to be lower.

This result indicates that the Llama 3.1-based model performs better with higher mapped subword coverage for the tokenizer size.
The results of Gemma 2 also demonstrate that unmapped tokens cause performance degradation.

\subsection{Token Efficiency}
Table~\ref{tab:num_subwords} shows that our approach reduces the number of subwords for low-resource languages compared to Llama 3.1, although \textsc{Focus} achieves an even greater reduction.
Our proposed method trains a dedicated tokenizer for the target languages compared with the Llama 3.1 tokenizer. As a result, our tokenization is more efficient than that of a multilingual tokenizer.
This result demonstrates that our approach can address token over-fragmentation, although performance improvement is limited to several languages.

\subsection{Multilingual Model vs. Monolingual Model as a Source Model}
Table~\ref{tab:result_ner} shows the micro F1 scores of RoBERTa-based models and XLM-R-based models.
This result indicates that our proposed method works better with a monolingual model for languages except Japanese and Sanskrit than with a multilingual model.
Our proposed method uses only English information to estimate the embeddings of target subwords. Thus, monolingual models are suitable.

\subsection{Language Characteristics}
A common characteristic among the languages that saw the most benefit from our approach---Uyghur, Khmer, and Manchu---is that they are genealogically distant from English.
Additionally, Uyghur and Khmer use different writing systems from English.
Although this difference can limit the RoBERTa and XLM-R performance, our proposed method improves performance.

However, Japanese is also genealogically distant from English and uses a different script, yet RoBERTa outperformed the proposed method.
A possible reason for this is the large number of character types in Japanese.
When a tokenizer is trained on limited data, the wide variety of characters can prevent subwords from achieving sufficiently robust representations. This limitation likely contributed to the constrained performance of our proposed method in Japanese.

Table~\ref{tab:result_ppl} indicates that our proposed method can handle languages with any script, thanks to language-specific tokenizers.

\subsection{Morphological Typological Features}
Our proposed method utilizes bilingual dictionaries, but these dictionaries generally only contain a single word form, excluding conjugated or inflected variants. 
This feature might also make it challenging to handle inflection or particles.
We analyze the results with respect to two morphological types: isolating-synthetic and agglutinative-fusional~\cite{whaley1997introduction}.

\paragraph{Synthesis.}
In languages used in the experiments, Khmer is isolating, German is relatively synthetic, and others are synthetic.
Since isolating languages do not have inflectional changes, and words are generally used in the form listed in the dictionaries,
such languages benefit from our proposed method shown in Table~\ref{tab:result_ner}.

\paragraph{Fusion.}
German, Old English, and Sanskrit have stronger fusional (inflected) characteristics, and Japanese, Uyghur, and Manchu have stronger agglutinative characteristics.
In fusional languages, affixes carry multiple grammatical functions, requiring more training data to capture these functions than in agglutinative languages, where each affix corresponds to a single grammatical function.
As a result, the performance improvement by our proposed method for German, Old English, and Sanskrit is limited, as shown in Table~\ref{tab:result_ner}.

\begin{table*}[t]
    \centering
    \begin{tabular}{lrrrrrrr}
    \toprule
         & German & Japanese & Old English & Uyghur & Sanskrit & Khmer & Manchu \\
    \midrule
        Ours (w/o LAPT) & 76.99  & 71.93 & 45.43 & 36.06 & 44.23 & 42.21 & 94.87 \\
        \hspace{1em}- Removal & 74.48 & 55.30 & 46.83 & 31.16 & 43.81 & 37.50 & 95.25 \\
        \hline
        Ours (w/ LAPT) & 76.43 & 73.60 & 52.71 &  64.52  & 42.08 & 62.96 & 92.87 \\
        \hspace{1em}- Removal & 59.96 & 29.70 & 29.75 & 49.89 & 53.70 & 57.27 & 41.58 \\
    \bottomrule
    \end{tabular}
    \caption{Micro F1 scores of MLMs for WikiANN and ManNER of RoBERTa-based models when skipping the (4) Removing step in our proposed method, finishing the subword mapping at the first iteration.}
    \label{tab:result_ner_ablation}
\end{table*}

\begin{table*}[!t]
    \centering
    \resizebox{\textwidth}{!}{
    \begin{tabular}{lrrrrrrr}
    \toprule
         & German & Japanese & Old English & Uyghur & Sanskrit & Khmer & Manchu \\
    \midrule
        Llama 3.1  \\
        Ours (w/o LAPT) & 444,876.62 & 473.72 & 46180.42 & 18,053.31 & 1,503.39 & 64,508.22 & 144,818.36 \\
        \hspace{1em}- Removal & 16,709.97 & 19.73 & 17,395.35 & 7,126.41 & 3,404.88 & 13,652.20 & 52,778.58 \\
        \hline
        Ours (w/ LAPT) & 88.61 & 4.42 & 406.53 & 168.43 & 3.56 & 4.32 & 502.02 \\
        \hspace{1em}- Removal & 25.00 & 2.98 & 28.25 & 55.62 & 3.07 & 3.67 & 549.36 \\
        \midrule
        Gemma 2  \\
        Ours (w/o LAPT) & 6,802,328.30 & 1,340.02 & 4,685,467.63 & 3,163,709.28 & 97,101.92 & 652,351.91 & 4,848,206.72 \\
        \hspace{1em}- Removal & 9,314.48 & 131.73 & 50,985.49 & 51,610.23 & 35,789.10 & 5,966.91 & 403,457.47 \\
        \hline
        Ours (w/ LAPT) & 595.14 & 5.83 & 1,324.49 & 83.32 & 10.57 & 16.80 & 509.64 \\
        \hspace{1em}- Removal & 18.98 & 5.98 & 37.22 & 1,650,481.13 & 6.91 & 3.71 & 1,238.87
 \\
    \bottomrule
    \end{tabular}
    }
    \caption{The perplexity when skipping the (4) Removing step in our proposed method and finishing the subword mapping at the first iteration.}
    \label{tab:result_ppl_ablation}
\end{table*}

\begin{figure}[t]
    \centering
    \begin{subfigure}{0.45\textwidth}
        \centering
        \includegraphics[width=\textwidth]{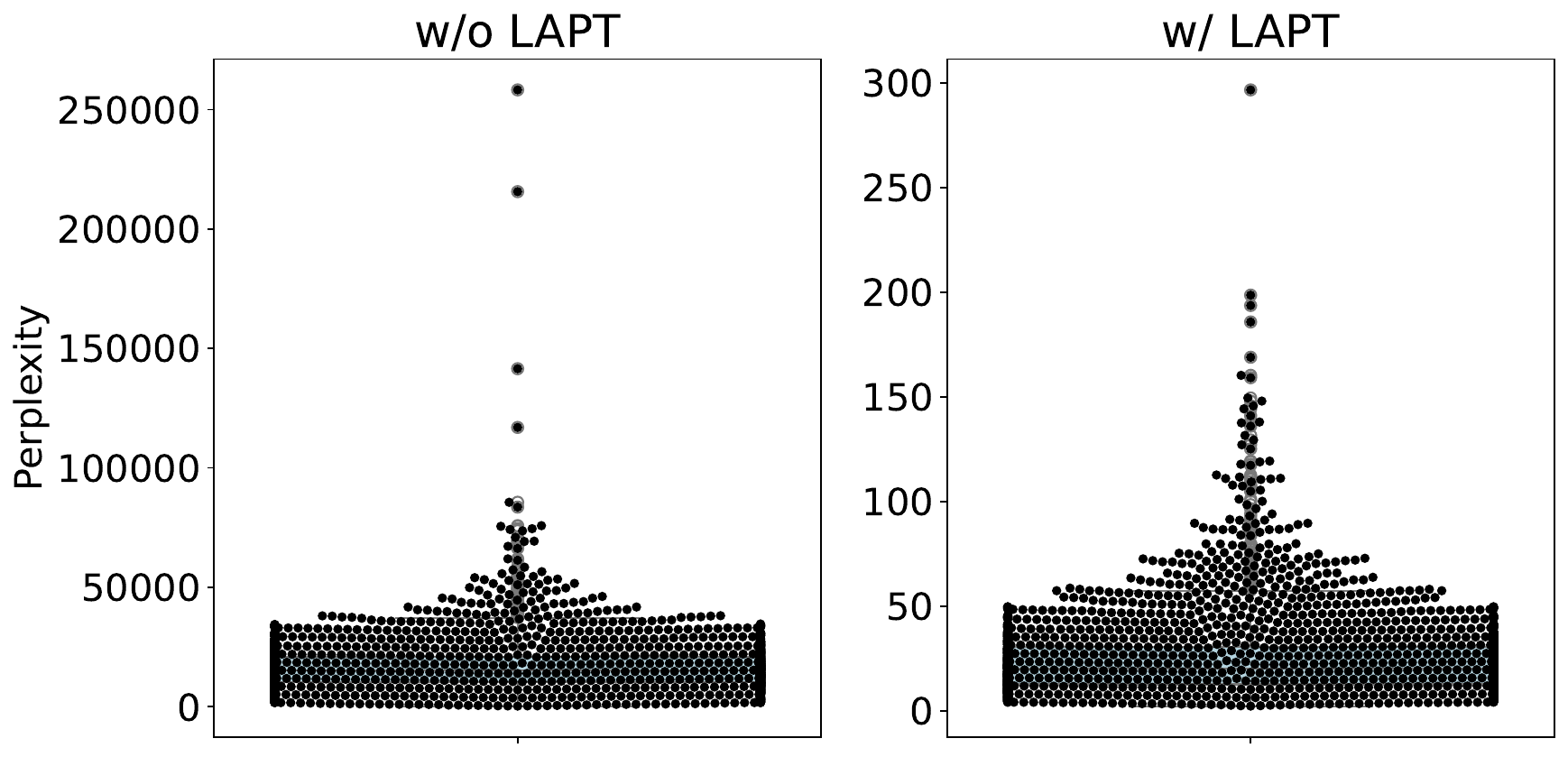}
        \caption{German}
    \end{subfigure}
    \hfill
    \begin{subfigure}{0.45\textwidth}
        \centering
        \includegraphics[width=\textwidth]{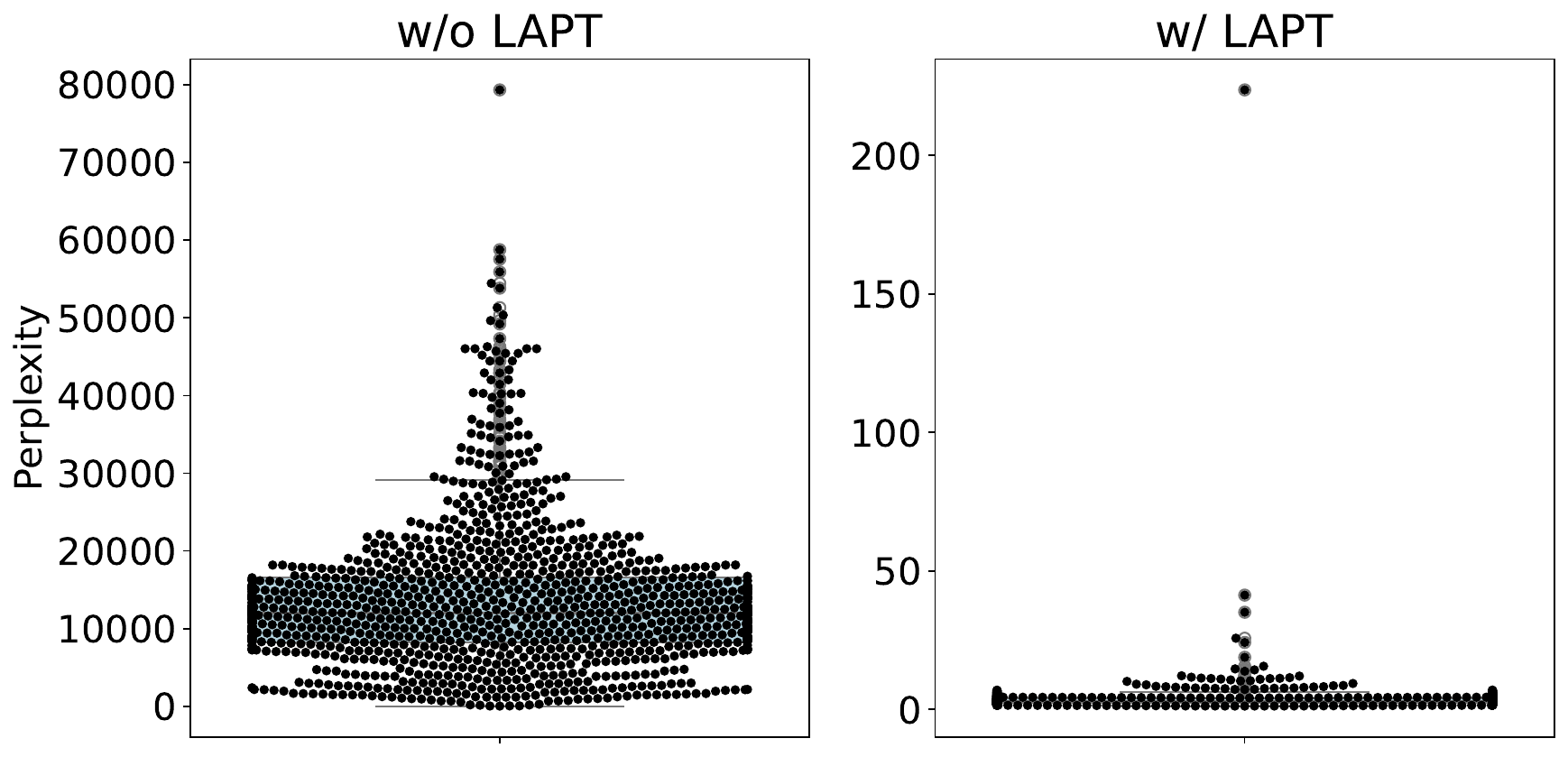}
        \caption{Khmer}
    \end{subfigure}
    \hfill
    \begin{subfigure}{0.45\textwidth}
        \centering
        \includegraphics[width=\textwidth]{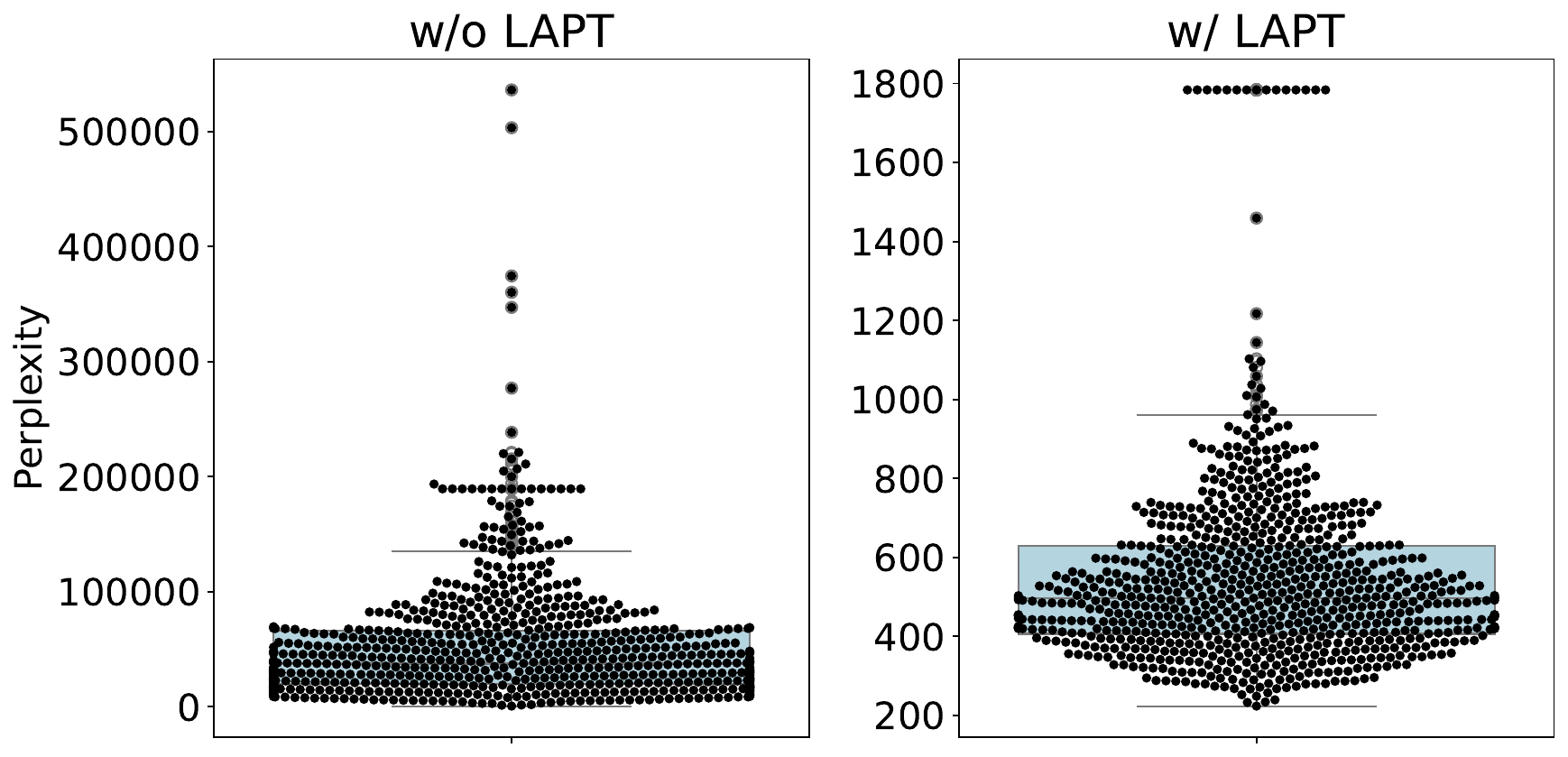}
        \caption{Manchu}
    \end{subfigure}
    \hfill
    \caption{The perplexity distribution and box plot of Llama 3.1-based our proposed method without the removal step.}
    \label{fig:wo_removal_ppl_distribution}
\end{figure}

\subsection{Ablation Study}
\paragraph{LAPT.}
LAPT improves NER performance for Japanese, Old English, Uyghur, and Khmer, while it has challenges for German, Sanskrit, and Manchu, as shown in Table~\ref{tab:result_ner}.
Considering that we fine-tune the models for NER, this suggests that LAPT plays an insignificant role.
This result is because the word order is insufficient for MLMs, and we also fine-tuned the model for NER.
On the other hand, LAPT improves the perplexity for all languages, as shown in Table~\ref{tab:result_ppl}.
This result suggests that LAPT tunes the model for target languages by incorporating word order, as the subword initialization does not consider word order.
On the other hand, Figure~\ref{fig:ppl_distribution} implies that there might be out-of-distribution samples with limited training data.

\paragraph{Mapped Subwords during Transferring}
Table~\ref{tab:mapped_iteration} shows the number of mapped subwords in each iteration.
The number remains almost the same across different models, indicating that iterative mapping is robust across base models' tokenizers for the same language.

\paragraph{Subword Removal.}
Table~\ref{tab:result_ner_ablation} and Table~\ref{tab:result_ppl_ablation} show the results of the ablation study about the mapped subword removal step in our proposed method.
Figure~\ref{fig:wo_removal_ppl_distribution} shows that OOV issues also occur without the removal.
The distribution of other languages and analysis are in Appendix~\ref{sec:appendix_ppl_distribution}.
These results indicate that the removal of mapped subwords and iteration play a crucial role in MLMs, specifically for LAPT.
On the other hand, the removal and iteration negatively affect CLMs, except in Japanese and Manchu.
The number of mapped subwords after the second loop varies across languages, as shown in Table~\ref{tab:mapped_iteration}.
For MLMs, the number of mapped subwords is essential, so the removal works well.
For example, approximately only 60\% of subwords in the vocabulary of the Japanese tokenizer are mapped in the first loop, which causes the performance decrease.
These results also align with the analysis in Section~\ref{sec:performance_ang_coverage}.
The exacerbation of perplexity indicates that the shorter subword mapping is noise for CLMs.
However, skipping the removal sometimes causes inefficient fine-tuning, as in Uyghur.
We conclude that removal can improve MLM performance and enhance the model's ability to handle extremely low-resource languages that are not included in the training data of pre-trained models.

\section{Conclusion}
In this study, we propose a dictionary-based cross-lingual vocabulary transfer approach.
This method utilizes bilingual dictionaries, which are available even for low-resource languages.
It iteratively maps a subword in a target language to subwords in a source language and initializes the target subword embeddings.
The experimental results demonstrate that our approach effectively initializes the subword embeddings for low-resource languages, indicating it is a promising approach for building language models for those languages.

\section*{Limitations}
\paragraph{Dictionaries and Languages.}
Dictionaries have only the unmarked form and not the inflected form in general, suggesting that handling inflection is challenging for our dictionary-based method.
Additionally, some online dictionaries (e.g., MUSE) suffer from quality issues~\cite{kementchedjhieva-etal-2019-lost}.
However, as this study demonstrates, the dictionary-based method performs better for low-resource languages than the existing method we used as a baseline.
This indicates that dictionaries are helpful for cross-lingual vocabulary transfer for low-resource languages despite the lack of grammatical information.

\paragraph{Language Set.}
In the experiments, we used seven languages: German, Japanese, Old English, Uyghur, Sanskrit, Khmer, and Manchu.
The number of selected languages is limited, and there are other languages with different linguistic features to test.
However, we include the language, Manchu, whose resources are less available on the Internet.
This means that we tested the language in which the training data of language models contains less data, which is valuable for low-resource languages.

\paragraph{Generalizability.}
Our proposed method shows challenges for several languages and might raise questions about the generalizability.
However, it is valuable if the method improves the performance of languages that face challenges with existing methods.

\paragraph{Evaluation.}
We evaluated CLMs by perplexity, which might raise questions about fair comparison when we compare a model with other models with different tokenizers.
The performance of downstream tasks can be used to evaluate CLMs~\cite{yamaguchi2024effectivelyexpandvocabularyllms, minixhofer-etal-2024-zeroshot, han2025adapters}.
However, we aim to apply our proposed method to low-resource languages, including extremely low-resource languages (e.g., Manchu), and the downstream tasks' datasets for CLMs are not available for those languages.
Thus, we use the perplexity normalized by word length, not subword length, for fair comparison.

\paragraph{Performance Improvement.}
Section~\ref{sec:result} shows that the performance improvement is limited with our proposed method for some languages.
There is room for performance improvement by refining the alignment method or hyperparameter tuning.
However, the limited improvement is a known problem in cross-lingual vocabulary transfer for low-resource languages~\cite{yamaguchi2024effectivelyexpandvocabularyllms}.
We aim to demonstrate the potential of a dictionary-based approach for cross-lingual vocabulary transfer in low-resource languages, and performance improvement will be the subject of a future study.

\paragraph{Model Expansion Instead of Replacement.}
In this study, we replace the source vocabulary with the target vocabulary instead of extending the source vocabulary.
Vocabulary expansion is sometimes needed for multilingualism.
However, vocabulary replacement performs better than expansion for target languages~\cite{dobler-de-melo-2023-focus, yamaguchi-etal-2024-empirical}, and sometimes performs better even in a source language~\cite{yamaguchi-etal-2024-empirical}.
In this study, we incorporate vocabulary replacement for this reason.

\paragraph{Tokenization.}
Since the trained tokenizer using dictionary entries has meaningful tokens in our proposed method, morpheme-aware tokenization~\cite{libovicky-helcl-2024-lexically, asgari2025morphbpemorphoawaretokenizerbridging} can be helpful.
Other tokenizers can also be used, such as SentencePiece~\cite{kudo-richardson-2018-sentencepiece}.
However, in this study, we trained BPE tokenizers, which are widely used tokenizer types, to compare with existing models.

\section*{Ethical Considerations}
\paragraph{Dataset.}
In this study, we use bilingual dictionaries and corpora from several sources.
We follow the given licenses for each dataset.
These datasets might contain identical or sensitive information, even though they are used widely.
The model's behavior in this study should be assessed before releasing it as an application. 

\paragraph{Use of AI Assistants.}
In this study, we have used GitHub Copilot for coding support.

\section*{Acknowledgement}
We thank the anonymous reviewers for their valuable comments and suggestions.
This project is partly supported by the Gemma Academic Program.

\bibliography{anthology,custom}

\begin{thebibliography}{43}
\providecommand{\natexlab}[1]{#1}

\bibitem[{Adams et~al.(2017)Adams, Makarucha, Neubig, Bird, and Cohn}]{adams-etal-2017-cross}
Oliver Adams, Adam Makarucha, Graham Neubig, Steven Bird, and Trevor Cohn. 2017.
\newblock \href {https://aclanthology.org/E17-1088/} {Cross-lingual word embeddings for low-resource language modeling}.
\newblock In \emph{Proceedings of the 15th Conference of the {E}uropean Chapter of the Association for Computational Linguistics: Volume 1, Long Papers}, pages 937--947, Valencia, Spain. Association for Computational Linguistics.

\bibitem[{Ahia et~al.(2023)Ahia, Kumar, Gonen, Kasai, Mortensen, Smith, and Tsvetkov}]{ahia-etal-2023-languages}
Orevaoghene Ahia, Sachin Kumar, Hila Gonen, Jungo Kasai, David Mortensen, Noah Smith, and Yulia Tsvetkov. 2023.
\newblock \href {https://doi.org/10.18653/v1/2023.emnlp-main.614} {Do all languages cost the same? tokenization in the era of commercial language models}.
\newblock In \emph{Proceedings of the 2023 Conference on Empirical Methods in Natural Language Processing}, pages 9904--9923, Singapore. Association for Computational Linguistics.

\bibitem[{Artetxe et~al.(2017)Artetxe, Labaka, and Agirre}]{artetxe-etal-2017-learning}
Mikel Artetxe, Gorka Labaka, and Eneko Agirre. 2017.
\newblock \href {https://doi.org/10.18653/v1/P17-1042} {Learning bilingual word embeddings with (almost) no bilingual data}.
\newblock In \emph{Proceedings of the 55th Annual Meeting of the Association for Computational Linguistics (Volume 1: Long Papers)}, pages 451--462, Vancouver, Canada. Association for Computational Linguistics.

\bibitem[{Asgari et~al.(2025)Asgari, Kheir, and Javaheri}]{asgari2025morphbpemorphoawaretokenizerbridging}
Ehsaneddin Asgari, Yassine~El Kheir, and Mohammad Ali~Sadraei Javaheri. 2025.
\newblock \href {https://arxiv.org/abs/2502.00894} {Morphbpe: A morpho-aware tokenizer bridging linguistic complexity for efficient llm training across morphologies}.
\newblock \emph{Preprint}, arXiv:2502.00894.

\bibitem[{Bojanowski et~al.(2017)Bojanowski, Grave, Joulin, and Mikolov}]{bojanowski-etal-2017-enriching}
Piotr Bojanowski, Edouard Grave, Armand Joulin, and Tomas Mikolov. 2017.
\newblock \href {https://doi.org/10.1162/tacl_a_00051} {Enriching word vectors with subword information}.
\newblock \emph{Transactions of the Association for Computational Linguistics}, 5:135--146.

\bibitem[{Brown et~al.(1993)Brown, Della~Pietra, Della~Pietra, and Mercer}]{brown-etal-1993-mathematics}
Peter~F. Brown, Stephen~A. Della~Pietra, Vincent~J. Della~Pietra, and Robert~L. Mercer. 1993.
\newblock \href {https://aclanthology.org/J93-2003/} {The mathematics of statistical machine translation: Parameter estimation}.
\newblock \emph{Computational Linguistics}, 19(2):263--311.

\bibitem[{Chau et~al.(2020)Chau, Lin, and Smith}]{chau-etal-2020-parsing}
Ethan~C. Chau, Lucy~H. Lin, and Noah~A. Smith. 2020.
\newblock \href {https://doi.org/10.18653/v1/2020.findings-emnlp.118} {Parsing with multilingual {BERT}, a small corpus, and a small treebank}.
\newblock In \emph{Findings of the Association for Computational Linguistics: EMNLP 2020}, pages 1324--1334, Online. Association for Computational Linguistics.

\bibitem[{Conneau et~al.(2020)Conneau, Khandelwal, Goyal, Chaudhary, Wenzek, Guzm{\'a}n, Grave, Ott, Zettlemoyer, and Stoyanov}]{conneau-etal-2020-unsupervised}
Alexis Conneau, Kartikay Khandelwal, Naman Goyal, Vishrav Chaudhary, Guillaume Wenzek, Francisco Guzm{\'a}n, Edouard Grave, Myle Ott, Luke Zettlemoyer, and Veselin Stoyanov. 2020.
\newblock \href {https://doi.org/10.18653/v1/2020.acl-main.747} {Unsupervised cross-lingual representation learning at scale}.
\newblock In \emph{Proceedings of the 58th Annual Meeting of the Association for Computational Linguistics}, pages 8440--8451, Online. Association for Computational Linguistics.

\bibitem[{Dobler and de~Melo(2023)}]{dobler-de-melo-2023-focus}
Konstantin Dobler and Gerard de~Melo. 2023.
\newblock \href {https://doi.org/10.18653/v1/2023.emnlp-main.829} {{FOCUS}: Effective embedding initialization for monolingual specialization of multilingual models}.
\newblock In \emph{Proceedings of the 2023 Conference on Empirical Methods in Natural Language Processing}, pages 13440--13454, Singapore. Association for Computational Linguistics.

\bibitem[{Dyer et~al.(2013)Dyer, Chahuneau, and Smith}]{dyer-etal-2013-simple}
Chris Dyer, Victor Chahuneau, and Noah~A. Smith. 2013.
\newblock \href {https://aclanthology.org/N13-1073/} {A simple, fast, and effective reparameterization of {IBM} model 2}.
\newblock In \emph{Proceedings of the 2013 Conference of the North {A}merican Chapter of the Association for Computational Linguistics: Human Language Technologies}, pages 644--648, Atlanta, Georgia. Association for Computational Linguistics.

\bibitem[{Fang and Cohn(2017)}]{fang-cohn-2017-model}
Meng Fang and Trevor Cohn. 2017.
\newblock \href {https://doi.org/10.18653/v1/P17-2093} {Model transfer for tagging low-resource languages using a bilingual dictionary}.
\newblock In \emph{Proceedings of the 55th Annual Meeting of the Association for Computational Linguistics (Volume 2: Short Papers)}, pages 587--593, Vancouver, Canada. Association for Computational Linguistics.

\bibitem[{Foundation()}]{wikidump}
Wikimedia Foundation.
\newblock \href {https://dumps.wikimedia.org} {Wikimedia downloads}.

\bibitem[{Gloeckle et~al.(2024)Gloeckle, Youbi~Idrissi, Roziere, Lopez-Paz, and Synnaeve}]{pmlr-v235-gloeckle24a}
Fabian Gloeckle, Badr Youbi~Idrissi, Baptiste Roziere, David Lopez-Paz, and Gabriel Synnaeve. 2024.
\newblock \href {https://proceedings.mlr.press/v235/gloeckle24a.html} {Better \& faster large language models via multi-token prediction}.
\newblock In \emph{Proceedings of the 41st International Conference on Machine Learning}, volume 235 of \emph{Proceedings of Machine Learning Research}, pages 15706--15734. PMLR.

\bibitem[{Grattafiori et~al.(2024)Grattafiori, Dubey, Jauhri, Pandey, Kadian, Al-Dahle, Letman, Mathur, Schelten, Vaughan, Yang, Fan, Goyal, Hartshorn, Yang, Mitra, Sravankumar, Korenev, Hinsvark, Rao, Zhang, Rodriguez, Gregerson, Spataru, Roziere, Biron, Tang, Chern, Caucheteux, Nayak, Bi, Marra, McConnell, Keller, Touret, Wu, Wong, Ferrer, Nikolaidis, Allonsius, Song, Pintz, Livshits, Wyatt, Esiobu, Choudhary, Mahajan, Garcia-Olano, Perino, Hupkes, Lakomkin, AlBadawy, Lobanova, Dinan, Smith, Radenovic, Guzmán, Zhang, Synnaeve, Lee, Anderson, Thattai, Nail, Mialon, Pang, Cucurell, Nguyen, Korevaar, Xu, Touvron, Zarov, Ibarra, Kloumann, Misra, Evtimov, Zhang, Copet, Lee, Geffert, Vranes, Park, Mahadeokar, Shah, van~der Linde, Billock, Hong, Lee, Fu, Chi, Huang, Liu, Wang, Yu, Bitton, Spisak, Park, Rocca, Johnstun, Saxe, Jia, Alwala, Prasad, Upasani, Plawiak, Li, Heafield, Stone, El-Arini, Iyer, Malik, Chiu, Bhalla, Lakhotia, Rantala-Yeary, van~der Maaten, Chen, Tan, Jenkins, Martin, Madaan, Malo, Blecher,
  Landzaat, de~Oliveira, Muzzi, Pasupuleti, Singh, Paluri, Kardas, Tsimpoukelli, Oldham, Rita, Pavlova, Kambadur, Lewis, Si, Singh, Hassan, Goyal, Torabi, Bashlykov, Bogoychev, Chatterji, Zhang, Duchenne, Çelebi, Alrassy, Zhang, Li, Vasic, Weng, Bhargava, Dubal, Krishnan, Koura, Xu, He, Dong, Srinivasan, Ganapathy, Calderer, Cabral, Stojnic, Raileanu, Maheswari, Girdhar, Patel, Sauvestre, Polidoro, Sumbaly, Taylor, Silva, Hou, Wang, Hosseini, Chennabasappa, Singh, Bell, Kim, Edunov, Nie, Narang, Raparthy, Shen, Wan, Bhosale, Zhang, Vandenhende, Batra, Whitman, Sootla, Collot, Gururangan, Borodinsky, Herman, Fowler, Sheasha, Georgiou, Scialom, Speckbacher, Mihaylov, Xiao, Karn, Goswami, Gupta, Ramanathan, Kerkez, Gonguet, Do, Vogeti, Albiero, Petrovic, Chu, Xiong, Fu, Meers, Martinet, Wang, Wang, Tan, Xia, Xie, Jia, Wang, Goldschlag, Gaur, Babaei, Wen, Song, Zhang, Li, Mao, Coudert, Yan, Chen, Papakipos, Singh, Srivastava, Jain, Kelsey, Shajnfeld, Gangidi, Victoria, Goldstand, Menon, Sharma, Boesenberg,
  Baevski, Feinstein, Kallet, Sangani, Teo, Yunus, Lupu, Alvarado, Caples, Gu, Ho, Poulton, Ryan, Ramchandani, Dong, Franco, Goyal, Saraf, Chowdhury, Gabriel, Bharambe, Eisenman, Yazdan, James, Maurer, Leonhardi, Huang, Loyd, Paola, Paranjape, Liu, Wu, Ni, Hancock, Wasti, Spence, Stojkovic, Gamido, Montalvo, Parker, Burton, Mejia, Liu, Wang, Kim, Zhou, Hu, Chu, Cai, Tindal, Feichtenhofer, Gao, Civin, Beaty, Kreymer, Li, Adkins, Xu, Testuggine, David, Parikh, Liskovich, Foss, Wang, Le, Holland, Dowling, Jamil, Montgomery, Presani, Hahn, Wood, Le, Brinkman, Arcaute, Dunbar, Smothers, Sun, Kreuk, Tian, Kokkinos, Ozgenel, Caggioni, Kanayet, Seide, Florez, Schwarz, Badeer, Swee, Halpern, Herman, Sizov, Guangyi, Zhang, Lakshminarayanan, Inan, Shojanazeri, Zou, Wang, Zha, Habeeb, Rudolph, Suk, Aspegren, Goldman, Zhan, Damlaj, Molybog, Tufanov, Leontiadis, Veliche, Gat, Weissman, Geboski, Kohli, Lam, Asher, Gaya, Marcus, Tang, Chan, Zhen, Reizenstein, Teboul, Zhong, Jin, Yang, Cummings, Carvill, Shepard, McPhie,
  Torres, Ginsburg, Wang, Wu, U, Saxena, Khandelwal, Zand, Matosich, Veeraraghavan, Michelena, Li, Jagadeesh, Huang, Chawla, Huang, Chen, Garg, A, Silva, Bell, Zhang, Guo, Yu, Moshkovich, Wehrstedt, Khabsa, Avalani, Bhatt, Mankus, Hasson, Lennie, Reso, Groshev, Naumov, Lathi, Keneally, Liu, Seltzer, Valko, Restrepo, Patel, Vyatskov, Samvelyan, Clark, Macey, Wang, Hermoso, Metanat, Rastegari, Bansal, Santhanam, Parks, White, Bawa, Singhal, Egebo, Usunier, Mehta, Laptev, Dong, Cheng, Chernoguz, Hart, Salpekar, Kalinli, Kent, Parekh, Saab, Balaji, Rittner, Bontrager, Roux, Dollar, Zvyagina, Ratanchandani, Yuvraj, Liang, Alao, Rodriguez, Ayub, Murthy, Nayani, Mitra, Parthasarathy, Li, Hogan, Battey, Wang, Howes, Rinott, Mehta, Siby, Bondu, Datta, Chugh, Hunt, Dhillon, Sidorov, Pan, Mahajan, Verma, Yamamoto, Ramaswamy, Lindsay, Lindsay, Feng, Lin, Zha, Patil, Shankar, Zhang, Zhang, Wang, Agarwal, Sajuyigbe, Chintala, Max, Chen, Kehoe, Satterfield, Govindaprasad, Gupta, Deng, Cho, Virk, Subramanian, Choudhury,
  Goldman, Remez, Glaser, Best, Koehler, Robinson, Li, Zhang, Matthews, Chou, Shaked, Vontimitta, Ajayi, Montanez, Mohan, Kumar, Mangla, Ionescu, Poenaru, Mihailescu, Ivanov, Li, Wang, Jiang, Bouaziz, Constable, Tang, Wu, Wang, Wu, Gao, Kleinman, Chen, Hu, Jia, Qi, Li, Zhang, Zhang, Adi, Nam, Yu, Wang, Zhao, Hao, Qian, Li, He, Rait, DeVito, Rosnbrick, Wen, Yang, Zhao, and Ma}]{grattafiori2024llama3herdmodels}
Aaron Grattafiori, Abhimanyu Dubey, Abhinav Jauhri, Abhinav Pandey, Abhishek Kadian, Ahmad Al-Dahle, Aiesha Letman, Akhil Mathur, Alan Schelten, Alex Vaughan, Amy Yang, Angela Fan, Anirudh Goyal, Anthony Hartshorn, Aobo Yang, Archi Mitra, Archie Sravankumar, Artem Korenev, Arthur Hinsvark, and 542 others. 2024.
\newblock \href {https://arxiv.org/abs/2407.21783} {The llama 3 herd of models}.
\newblock \emph{Preprint}, arXiv:2407.21783.

\bibitem[{Han et~al.(2025)Han, Eriguchi, Xu, Hoang, Carpuat, and Khayrallah}]{han2025adapters}
HyoJung Han, Akiko Eriguchi, Haoran Xu, Hieu Hoang, Marine Carpuat, and Huda Khayrallah. 2025.
\newblock \href {https://openreview.net/forum?id=KxQRHOre9D} {Adapters for altering {LLM} vocabularies: What languages benefit the most?}
\newblock In \emph{The Thirteenth International Conference on Learning Representations}.

\bibitem[{Hu et~al.(2022)Hu, yelong shen, Wallis, Allen-Zhu, Li, Wang, Wang, and Chen}]{hu2022lora}
Edward~J Hu, yelong shen, Phillip Wallis, Zeyuan Allen-Zhu, Yuanzhi Li, Shean Wang, Lu~Wang, and Weizhu Chen. 2022.
\newblock \href {https://openreview.net/forum?id=nZeVKeeFYf9} {Lo{RA}: Low-rank adaptation of large language models}.
\newblock In \emph{International Conference on Learning Representations}.

\bibitem[{Kementchedjhieva et~al.(2019)Kementchedjhieva, Hartmann, and S{\o}gaard}]{kementchedjhieva-etal-2019-lost}
Yova Kementchedjhieva, Mareike Hartmann, and Anders S{\o}gaard. 2019.
\newblock \href {https://doi.org/10.18653/v1/D19-1328} {Lost in evaluation: Misleading benchmarks for bilingual dictionary induction}.
\newblock In \emph{Proceedings of the 2019 Conference on Empirical Methods in Natural Language Processing and the 9th International Joint Conference on Natural Language Processing (EMNLP-IJCNLP)}, pages 3336--3341, Hong Kong, China. Association for Computational Linguistics.

\bibitem[{Kudo and Richardson(2018)}]{kudo-richardson-2018-sentencepiece}
Taku Kudo and John Richardson. 2018.
\newblock \href {https://doi.org/10.18653/v1/D18-2012} {{S}entence{P}iece: A simple and language independent subword tokenizer and detokenizer for neural text processing}.
\newblock In \emph{Proceedings of the 2018 Conference on Empirical Methods in Natural Language Processing: System Demonstrations}, pages 66--71, Brussels, Belgium. Association for Computational Linguistics.

\bibitem[{Kudo et~al.(2004)Kudo, Yamamoto, and Matsumoto}]{kudo-etal-2004-applying}
Taku Kudo, Kaoru Yamamoto, and Yuji Matsumoto. 2004.
\newblock \href {https://aclanthology.org/W04-3230/} {Applying conditional random fields to {J}apanese morphological analysis}.
\newblock In \emph{Proceedings of the 2004 Conference on Empirical Methods in Natural Language Processing}, pages 230--237, Barcelona, Spain. Association for Computational Linguistics.

\bibitem[{Lample et~al.(2017)Lample, Conneau, Denoyer, and Ranzato}]{lample2017unsupervised}
Guillaume Lample, Alexis Conneau, Ludovic Denoyer, and Marc'Aurelio Ranzato. 2017.
\newblock \href {https://arxiv.org/abs/1711.00043} {Unsupervised machine translation using monolingual corpora only}.
\newblock \emph{Preprint}, arXiv:1711.00043.

\bibitem[{Lee et~al.(2024)Lee, Byun, Seo, and Kang}]{lee-etal-2024-manner}
Sangah Lee, Sungjoo Byun, Jean Seo, and Minha Kang. 2024.
\newblock \href {https://aclanthology.org/2024.lrec-main.961/} {{M}an{NER} {\&} {M}an{POS}: Pioneering {NLP} for endangered {M}anchu language}.
\newblock In \emph{Proceedings of the 2024 Joint International Conference on Computational Linguistics, Language Resources and Evaluation (LREC-COLING 2024)}, pages 11030--11039, Torino, Italia. ELRA and ICCL.

\bibitem[{Libovick{\'y} and Helcl(2024)}]{libovicky-helcl-2024-lexically}
Jind{\v{r}}ich Libovick{\'y} and Jind{\v{r}}ich Helcl. 2024.
\newblock \href {https://doi.org/10.18653/v1/2024.emnlp-main.421} {Lexically grounded subword segmentation}.
\newblock In \emph{Proceedings of the 2024 Conference on Empirical Methods in Natural Language Processing}, pages 7403--7420, Miami, Florida, USA. Association for Computational Linguistics.

\bibitem[{Liu et~al.(2024)Liu, Lin, Wang, and Schuetze}]{liu-etal-2024-ofa}
Yihong Liu, Peiqin Lin, Mingyang Wang, and Hinrich Schuetze. 2024.
\newblock \href {https://doi.org/10.18653/v1/2024.findings-naacl.68} {{OFA}: A framework of initializing unseen subword embeddings for efficient large-scale multilingual continued pretraining}.
\newblock In \emph{Findings of the Association for Computational Linguistics: NAACL 2024}, pages 1067--1097, Mexico City, Mexico. Association for Computational Linguistics.

\bibitem[{Liu et~al.(2019)Liu, Ott, Goyal, Du, Joshi, Chen, Levy, Lewis, Zettlemoyer, and Stoyanov}]{liu2019robertarobustlyoptimizedbert}
Yinhan Liu, Myle Ott, Naman Goyal, Jingfei Du, Mandar Joshi, Danqi Chen, Omer Levy, Mike Lewis, Luke Zettlemoyer, and Veselin Stoyanov. 2019.
\newblock \href {https://arxiv.org/abs/1907.11692} {Ro{BERT}a: A robustly optimized bert pretraining approach}.
\newblock \emph{Preprint}, arXiv:1907.11692.

\bibitem[{Minixhofer et~al.(2022)Minixhofer, Paischer, and Rekabsaz}]{minixhofer-etal-2022-wechsel}
Benjamin Minixhofer, Fabian Paischer, and Navid Rekabsaz. 2022.
\newblock \href {https://doi.org/10.18653/v1/2022.naacl-main.293} {{WECHSEL}: Effective initialization of subword embeddings for cross-lingual transfer of monolingual language models}.
\newblock In \emph{Proceedings of the 2022 Conference of the North American Chapter of the Association for Computational Linguistics: Human Language Technologies}, pages 3992--4006, Seattle, United States. Association for Computational Linguistics.

\bibitem[{Minixhofer et~al.(2024)Minixhofer, Ponti, and Vuli\'{c}}]{minixhofer-etal-2024-zeroshot}
Benjamin Minixhofer, Edoardo~M. Ponti, and Ivan Vuli\'{c}. 2024.
\newblock \href {https://proceedings.neurips.cc/paper_files/paper/2024/file/532ce4fcf853023c4cf2ac38cbc5d002-Paper-Conference.pdf} {Zero-shot tokenizer transfer}.
\newblock In \emph{Advances in Neural Information Processing Systems}, volume~37, pages 46791--46818. Curran Associates, Inc.

\bibitem[{Moroni et~al.(2025)Moroni, Puccetti, Huguet~Cabot, Bejgu, Miaschi, Barba, Dell{'}Orletta, Esuli, and Navigli}]{moroni-etal-2025-optimizing}
Luca Moroni, Giovanni Puccetti, Pere-Llu{\'i}s Huguet~Cabot, Andrei~Stefan Bejgu, Alessio Miaschi, Edoardo Barba, Felice Dell{'}Orletta, Andrea Esuli, and Roberto Navigli. 2025.
\newblock \href {https://aclanthology.org/2025.findings-naacl.371/} {Optimizing {LLM}s for {I}talian: Reducing token fertility and enhancing efficiency through vocabulary adaptation}.
\newblock In \emph{Findings of the Association for Computational Linguistics: NAACL 2025}, pages 6646--6660, Albuquerque, New Mexico. Association for Computational Linguistics.

\bibitem[{Norman(2013)}]{Norman2013-tq}
Jerry Norman. 2013.
\newblock \emph{A comprehensive {Manchu-English} dictionary}.
\newblock Harvard-Yenching Institute Monograph Series. Harvard University, Asia Center, Cambridge, MA.

\bibitem[{Pan et~al.(2017)Pan, Zhang, May, Nothman, Knight, and Ji}]{pan-etal-2017-cross}
Xiaoman Pan, Boliang Zhang, Jonathan May, Joel Nothman, Kevin Knight, and Heng Ji. 2017.
\newblock \href {https://doi.org/10.18653/v1/P17-1178} {Cross-lingual name tagging and linking for 282 languages}.
\newblock In \emph{Proceedings of the 55th Annual Meeting of the Association for Computational Linguistics (Volume 1: Long Papers)}, pages 1946--1958, Vancouver, Canada. Association for Computational Linguistics.

\bibitem[{Petrov et~al.(2023)Petrov, Malfa, Torr, and Bibi}]{petrov2023language}
Aleksandar Petrov, Emanuele~La Malfa, Philip Torr, and Adel Bibi. 2023.
\newblock \href {https://openreview.net/forum?id=78yDLKi95p} {Language model tokenizers introduce unfairness between languages}.
\newblock In \emph{Thirty-seventh Conference on Neural Information Processing Systems}.

\bibitem[{Pham et~al.(2024)Pham, Le, and Luu}]{pham-etal-2024-unibridge}
Trinh Pham, Khoi Le, and Anh~Tuan Luu. 2024.
\newblock \href {https://doi.org/10.18653/v1/2024.acl-long.174} {{U}ni{B}ridge: A unified approach to cross-lingual transfer learning for low-resource languages}.
\newblock In \emph{Proceedings of the 62nd Annual Meeting of the Association for Computational Linguistics (Volume 1: Long Papers)}, pages 3168--3184, Bangkok, Thailand. Association for Computational Linguistics.

\bibitem[{Rahimi et~al.(2019)Rahimi, Li, and Cohn}]{rahimi-etal-2019-massively}
Afshin Rahimi, Yuan Li, and Trevor Cohn. 2019.
\newblock \href {https://doi.org/10.18653/v1/P19-1015} {Massively multilingual transfer for {NER}}.
\newblock In \emph{Proceedings of the 57th Annual Meeting of the Association for Computational Linguistics}, pages 151--164, Florence, Italy. Association for Computational Linguistics.

\bibitem[{Remy et~al.(2024)Remy, Delobelle, Avetisyan, Khabibullina, de~Lhoneux, and Demeester}]{remy-delobelle2024transtokenization}
Fran{\c{c}}ois Remy, Pieter Delobelle, Hayastan Avetisyan, Alfiya Khabibullina, Miryam de~Lhoneux, and Thomas Demeester. 2024.
\newblock \href {https://openreview.net/forum?id=sBxvoDhvao} {Trans-tokenization and cross-lingual vocabulary transfers: Language adaptation of {LLM}s for low-resource {NLP}}.
\newblock In \emph{First Conference on Language Modeling}.

\bibitem[{Roh et~al.(2020)Roh, Oh, and Lee}]{roh2020unigramnormalizedperplexitylanguagemodel}
Jihyeon Roh, Sang-Hoon Oh, and Soo-Young Lee. 2020.
\newblock \href {https://arxiv.org/abs/2011.13220} {Unigram-normalized perplexity as a language model performance measure with different vocabulary sizes}.
\newblock \emph{Preprint}, arXiv:2011.13220.

\bibitem[{Sennrich et~al.(2016)Sennrich, Haddow, and Birch}]{sennrich-etal-2016-neural}
Rico Sennrich, Barry Haddow, and Alexandra Birch. 2016.
\newblock \href {https://doi.org/10.18653/v1/P16-1162} {Neural machine translation of rare words with subword units}.
\newblock In \emph{Proceedings of the 54th Annual Meeting of the Association for Computational Linguistics (Volume 1: Long Papers)}, pages 1715--1725, Berlin, Germany. Association for Computational Linguistics.

\bibitem[{Team et~al.(2024)Team, Riviere, Pathak, Sessa, Hardin, Bhupatiraju, Hussenot, Mesnard, Shahriari, Ramé, Ferret, Liu, Tafti, Friesen, Casbon, Ramos, Kumar, Lan, Jerome, Tsitsulin, Vieillard, Stanczyk, Girgin, Momchev, Hoffman, Thakoor, Grill, Neyshabur, Bachem, Walton, Severyn, Parrish, Ahmad, Hutchison, Abdagic, Carl, Shen, Brock, Coenen, Laforge, Paterson, Bastian, Piot, Wu, Royal, Chen, Kumar, Perry, Welty, Choquette-Choo, Sinopalnikov, Weinberger, Vijaykumar, Rogozińska, Herbison, Bandy, Wang, Noland, Moreira, Senter, Eltyshev, Visin, Rasskin, Wei, Cameron, Martins, Hashemi, Klimczak-Plucińska, Batra, Dhand, Nardini, Mein, Zhou, Svensson, Stanway, Chan, Zhou, Carrasqueira, Iljazi, Becker, Fernandez, van Amersfoort, Gordon, Lipschultz, Newlan, yeong Ji, Mohamed, Badola, Black, Millican, McDonell, Nguyen, Sodhia, Greene, Sjoesund, Usui, Sifre, Heuermann, Lago, McNealus, Soares, Kilpatrick, Dixon, Martins, Reid, Singh, Iverson, Görner, Velloso, Wirth, Davidow, Miller, Rahtz, Watson, Risdal,
  Kazemi, Moynihan, Zhang, Kahng, Park, Rahman, Khatwani, Dao, Bardoliwalla, Devanathan, Dumai, Chauhan, Wahltinez, Botarda, Barnes, Barham, Michel, Jin, Georgiev, Culliton, Kuppala, Comanescu, Merhej, Jana, Rokni, Agarwal, Mullins, Saadat, Carthy, Cogan, Perrin, Arnold, Krause, Dai, Garg, Sheth, Ronstrom, Chan, Jordan, Yu, Eccles, Hennigan, Kocisky, Doshi, Jain, Yadav, Meshram, Dharmadhikari, Barkley, Wei, Ye, Han, Kwon, Xu, Shen, Gong, Wei, Cotruta, Kirk, Rao, Giang, Peran, Warkentin, Collins, Barral, Ghahramani, Hadsell, Sculley, Banks, Dragan, Petrov, Vinyals, Dean, Hassabis, Kavukcuoglu, Farabet, Buchatskaya, Borgeaud, Fiedel, Joulin, Kenealy, Dadashi, and Andreev}]{gemmateam2024gemma2improvingopen}
Gemma Team, Morgane Riviere, Shreya Pathak, Pier~Giuseppe Sessa, Cassidy Hardin, Surya Bhupatiraju, Léonard Hussenot, Thomas Mesnard, Bobak Shahriari, Alexandre Ramé, Johan Ferret, Peter Liu, Pouya Tafti, Abe Friesen, Michelle Casbon, Sabela Ramos, Ravin Kumar, Charline~Le Lan, Sammy Jerome, and 179 others. 2024.
\newblock \href {https://arxiv.org/abs/2408.00118} {Gemma 2: Improving open language models at a practical size}.
\newblock \emph{Preprint}, arXiv:2408.00118.

\bibitem[{van~der Maaten and Hinton(2008)}]{JMLR:v9:vandermaaten08a}
Laurens van~der Maaten and Geoffrey Hinton. 2008.
\newblock \href {http://jmlr.org/papers/v9/vandermaaten08a.html} {Visualizing data using t-sne}.
\newblock \emph{Journal of Machine Learning Research}, 9(86):2579--2605.

\bibitem[{Vasselli et~al.(2025)Vasselli, Sakajo, Mart{\'i}nez~Peguero, Hudi, and Watanabe}]{vasselli-etal-2025-leveraging}
Justin Vasselli, Haruki Sakajo, Arturo Mart{\'i}nez~Peguero, Frederikus Hudi, and Taro Watanabe. 2025.
\newblock \href {https://aclanthology.org/2025.americasnlp-1.13/} {Leveraging dictionaries and grammar rules for the creation of educational materials for indigenous languages}.
\newblock In \emph{Proceedings of the Fifth Workshop on NLP for Indigenous Languages of the Americas (AmericasNLP)}, pages 112--118, Albuquerque, New Mexico. Association for Computational Linguistics.

\bibitem[{Wang et~al.(2020)Wang, K, Mayhew, and Roth}]{wang-etal-2020-extending}
Zihan Wang, Karthikeyan K, Stephen Mayhew, and Dan Roth. 2020.
\newblock \href {https://doi.org/10.18653/v1/2020.findings-emnlp.240} {Extending multilingual {BERT} to low-resource languages}.
\newblock In \emph{Findings of the Association for Computational Linguistics: EMNLP 2020}, pages 2649--2656, Online. Association for Computational Linguistics.

\bibitem[{Whaley(1997)}]{whaley1997introduction}
L.J. Whaley. 1997.
\newblock \href {https://books.google.co.jp/books?id=S7oRYzV5SJgC} {\emph{Introduction to Typology: The Unity and Diversity of Language}}.
\newblock SAGE Publications.

\bibitem[{Wolf et~al.(2020)Wolf, Debut, Sanh, Chaumond, Delangue, Moi, Cistac, Rault, Louf, Funtowicz, Davison, Shleifer, von Platen, Ma, Jernite, Plu, Xu, Le~Scao, Gugger, Drame, Lhoest, and Rush}]{wolf-etal-2020-transformers}
Thomas Wolf, Lysandre Debut, Victor Sanh, Julien Chaumond, Clement Delangue, Anthony Moi, Pierric Cistac, Tim Rault, Remi Louf, Morgan Funtowicz, Joe Davison, Sam Shleifer, Patrick von Platen, Clara Ma, Yacine Jernite, Julien Plu, Canwen Xu, Teven Le~Scao, Sylvain Gugger, and 3 others. 2020.
\newblock \href {https://doi.org/10.18653/v1/2020.emnlp-demos.6} {Transformers: State-of-the-art natural language processing}.
\newblock In \emph{Proceedings of the 2020 Conference on Empirical Methods in Natural Language Processing: System Demonstrations}, pages 38--45, Online. Association for Computational Linguistics.

\bibitem[{Yamaguchi et~al.(2024{\natexlab{a}})Yamaguchi, Villavicencio, and Aletras}]{yamaguchi-etal-2024-empirical}
Atsuki Yamaguchi, Aline Villavicencio, and Nikolaos Aletras. 2024{\natexlab{a}}.
\newblock \href {https://doi.org/10.18653/v1/2024.findings-emnlp.396} {An empirical study on cross-lingual vocabulary adaptation for efficient language model inference}.
\newblock In \emph{Findings of the Association for Computational Linguistics: EMNLP 2024}, pages 6760--6785, Miami, Florida, USA. Association for Computational Linguistics.

\bibitem[{Yamaguchi et~al.(2024{\natexlab{b}})Yamaguchi, Villavicencio, and Aletras}]{yamaguchi2024effectivelyexpandvocabularyllms}
Atsuki Yamaguchi, Aline Villavicencio, and Nikolaos Aletras. 2024{\natexlab{b}}.
\newblock \href {https://arxiv.org/abs/2406.11477} {How can we effectively expand the vocabulary of llms with 0.01gb of target language text?}
\newblock \emph{Preprint}, arXiv:2406.11477.

\end{thebibliography}

\appendix

\section{Details of Experimental Setups}
\label{sec:detai_experimental_setups}
We use the Hugging Face Transformers library~\cite{wolf-etal-2020-transformers} to utilize the language models. Table~\ref{tab:models} lists the models used in this study along with their corresponding Hugging Face IDs.
The hyperparameters used in the experiments are in Tables~\ref{tab:params_lapt}~and~\ref{tab:params_finetuning}.
We execute the EM algorithm seven times with fast\_align.
We use a single NVIDIA GeForce RTX 3090 GPU for training MLMs and inference of MLMs and CLMs.
We also use eight NVIDIA L4 GPUs, a single NVIDIA A100-SXM4-40GB GPU, or one or two NVIDIA RTX A6000 or NVIDIA RTX 6000 Ada Generation GPUs for training of CLMs.

\section{Details of the Results}
\label{sec:appendix_results_detail}

\subsection{Visualization}
\label{sec:appendix_visualization}

\begin{figure}[!t]
    \centering
    \includegraphics[width=0.5\linewidth]{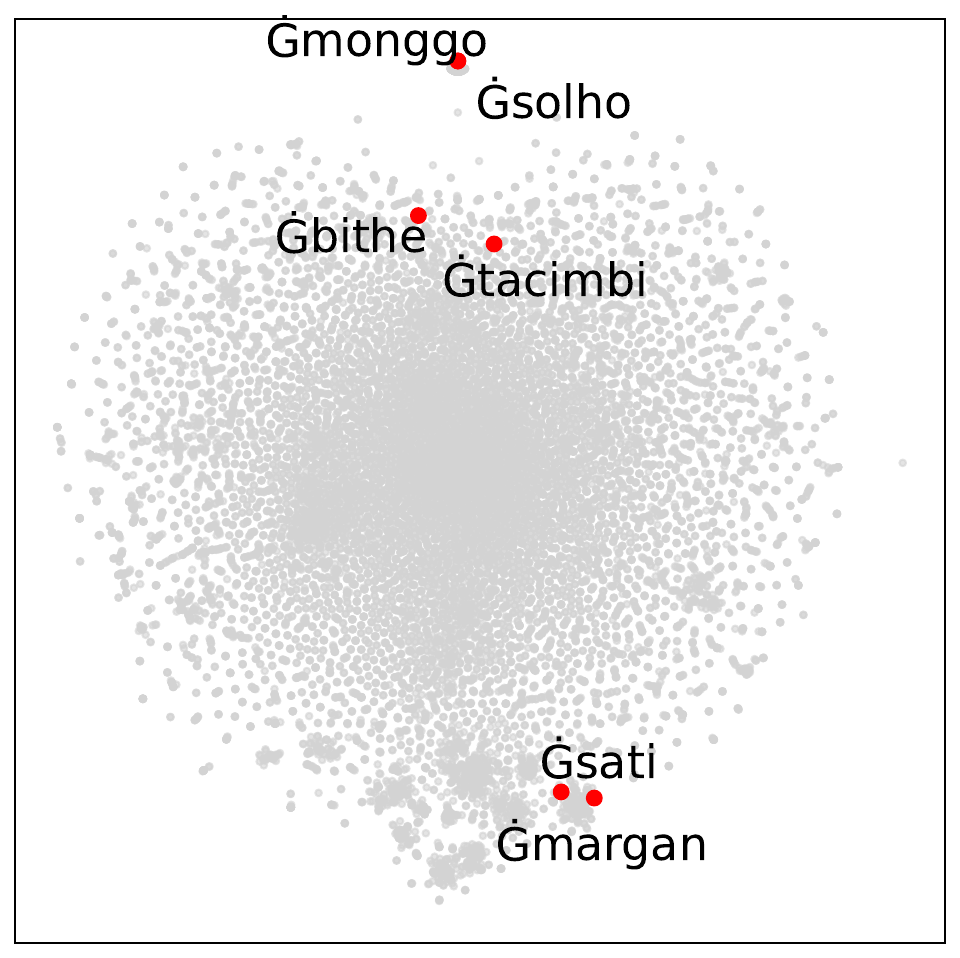}
    \caption{The subword embedding visualization with t-SNE for Manchu after transferring from RoBERTa without LAPT. ``monggo'', ``slho'', ``bithe'', ``tacimbi'', ``sati'', ``margan'' mean ``Mongol'', ``Korea'', ``book'', ``lern'', ``bear'', and ``deer'' respectively.}
    \label{fig:embedding_visualization}
\end{figure}

Figure~\ref{fig:embedding_visualization} visualizes the subword embedding with t-SNE~\cite{JMLR:v9:vandermaaten08a} for Manchu, emphasizing several subwords' points.
This figure demonstrates that the subwords that have similar meanings or concepts to each other are positioned close to each other.

\subsection{Removal and Perplexity Distribution}
\label{sec:appendix_ppl_distribution}
\begin{figure*}
    \centering
    \begin{subfigure}{0.3\textwidth}
        \centering
        \includegraphics[width=\textwidth]{assets/ppl_llama_de.pdf}
        \caption{German}
    \end{subfigure}
    \hfill
    \begin{subfigure}{0.3\textwidth}
        \centering
        \includegraphics[width=\textwidth]{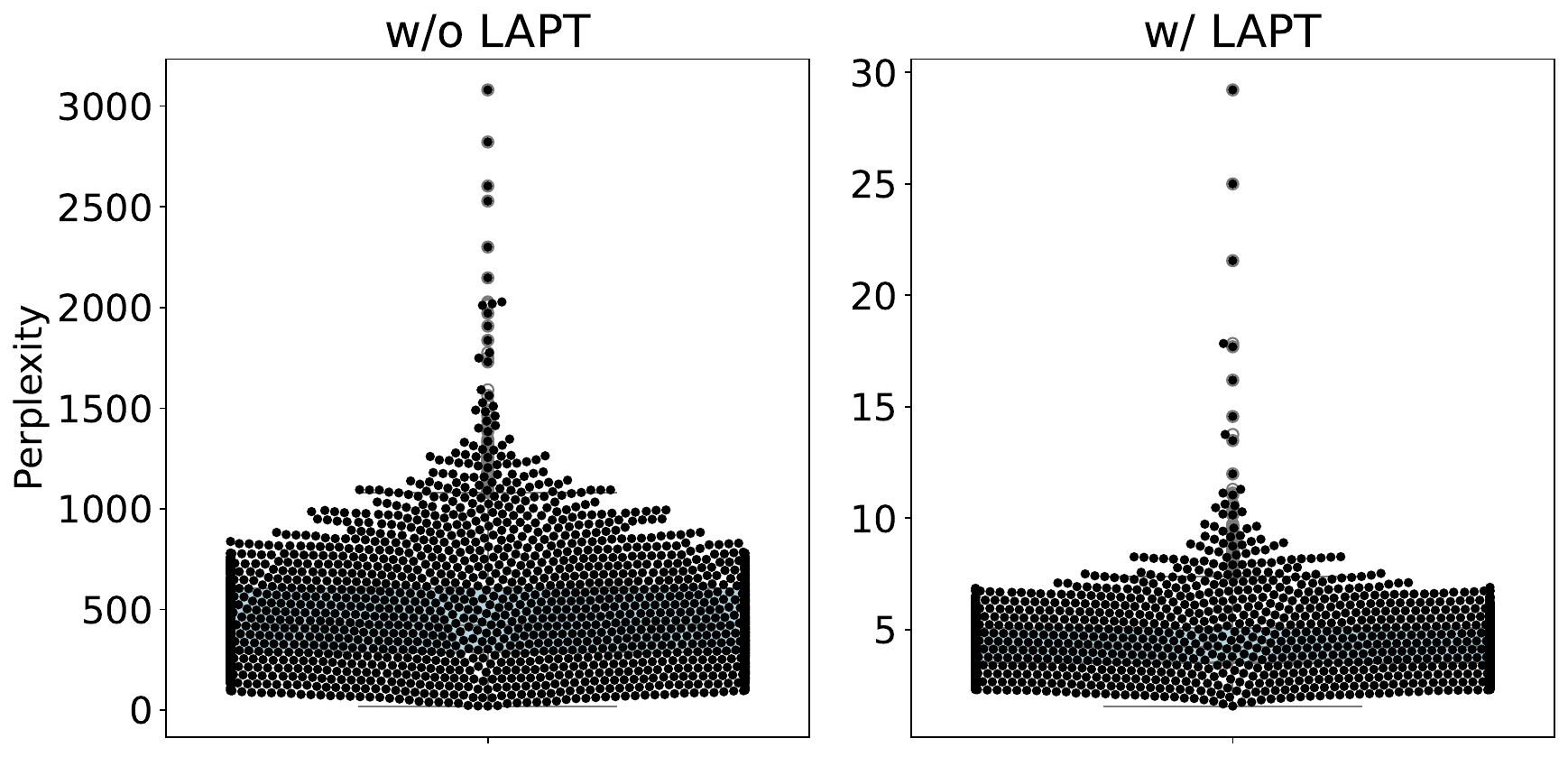}
        \caption{Japanese}
    \end{subfigure}
    \hfill
    \begin{subfigure}{0.3\textwidth}
        \centering
        \includegraphics[width=\textwidth]{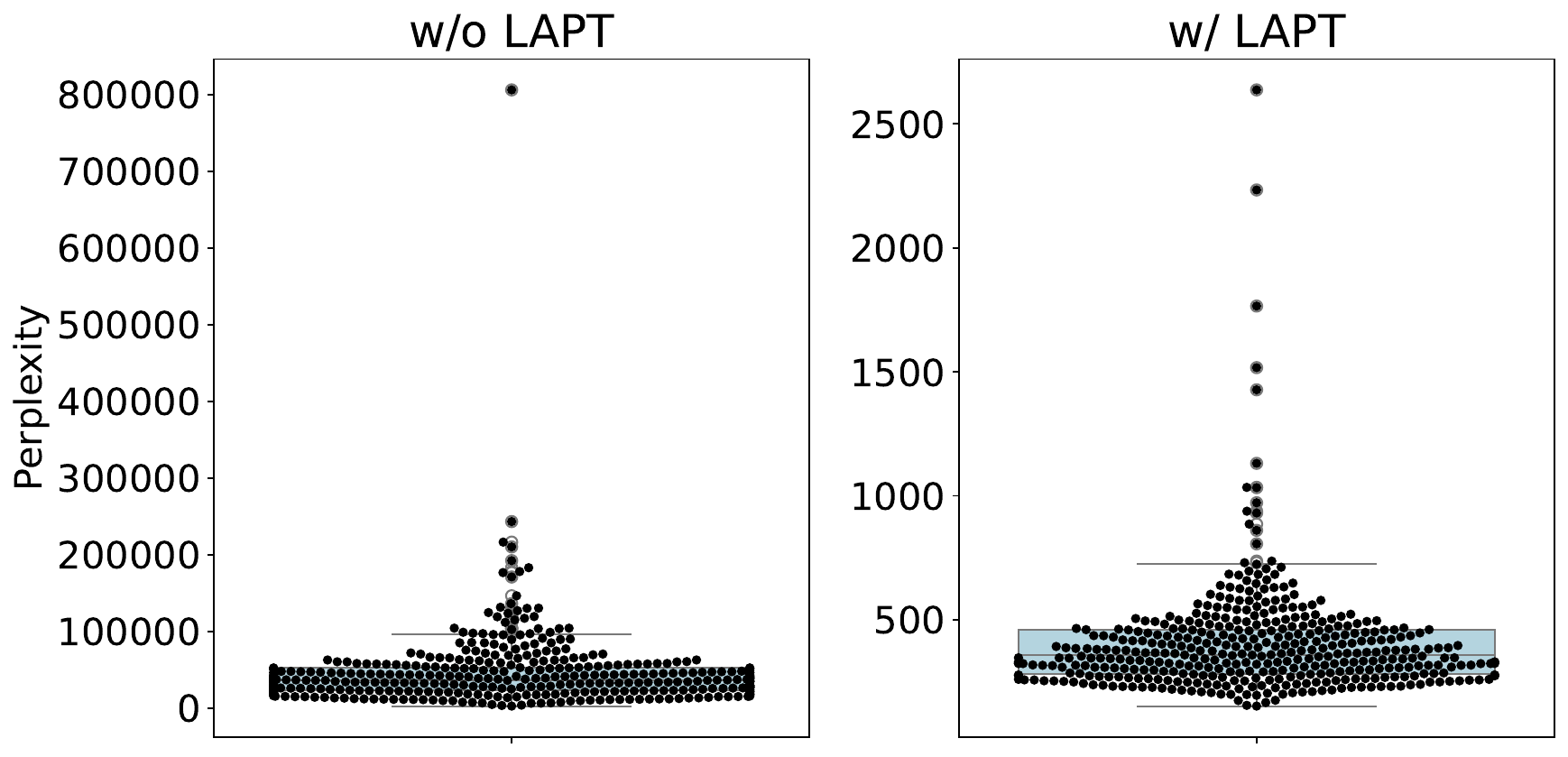}
        \caption{Old English}
    \end{subfigure}
    \hfill
    \begin{subfigure}{0.3\textwidth}
        \centering
        \includegraphics[width=\textwidth]{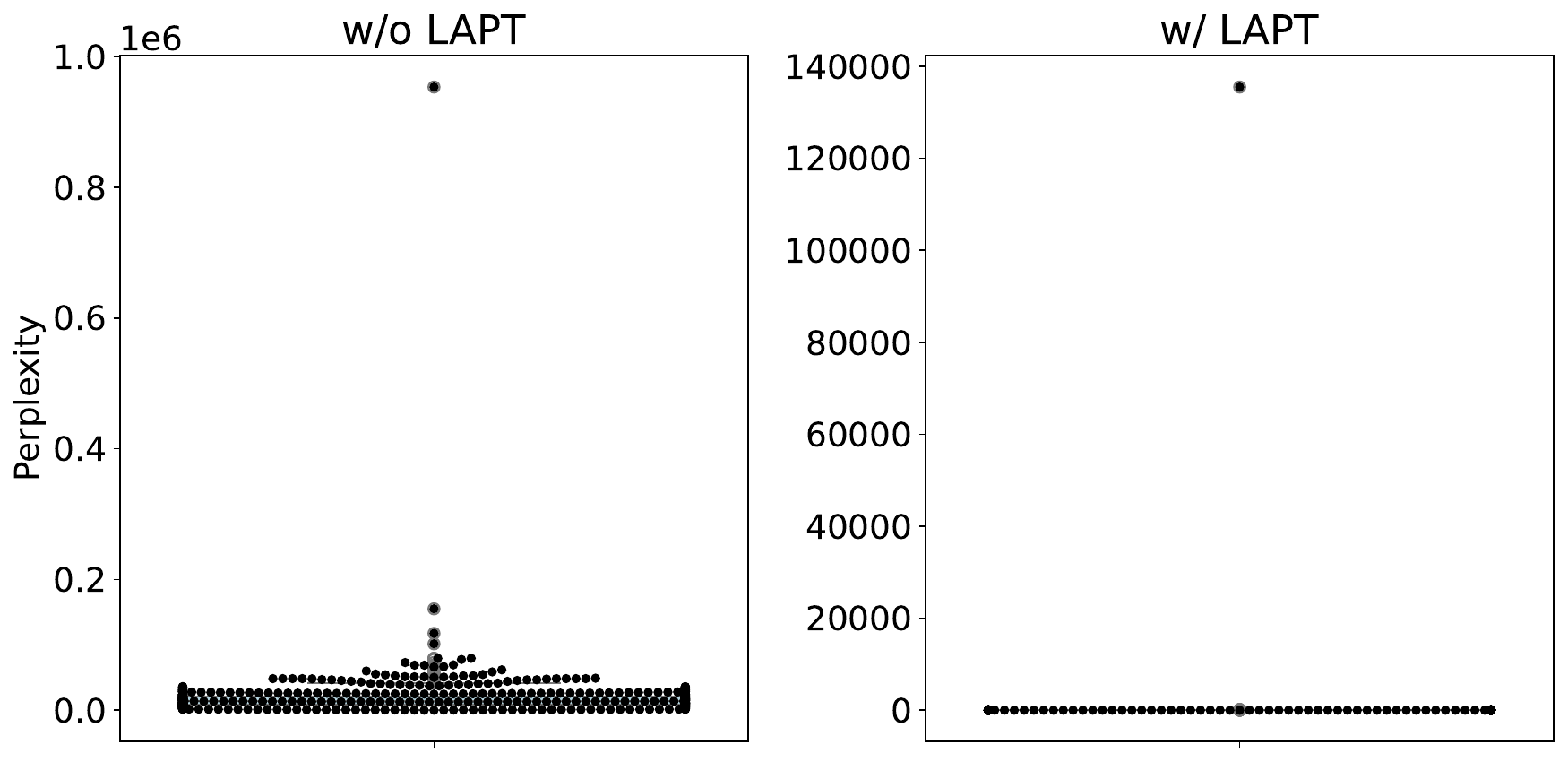}
        \caption{Uyghur}
    \end{subfigure}
    \hfill
    \begin{subfigure}{0.3\textwidth}
        \centering
        \includegraphics[width=\textwidth]{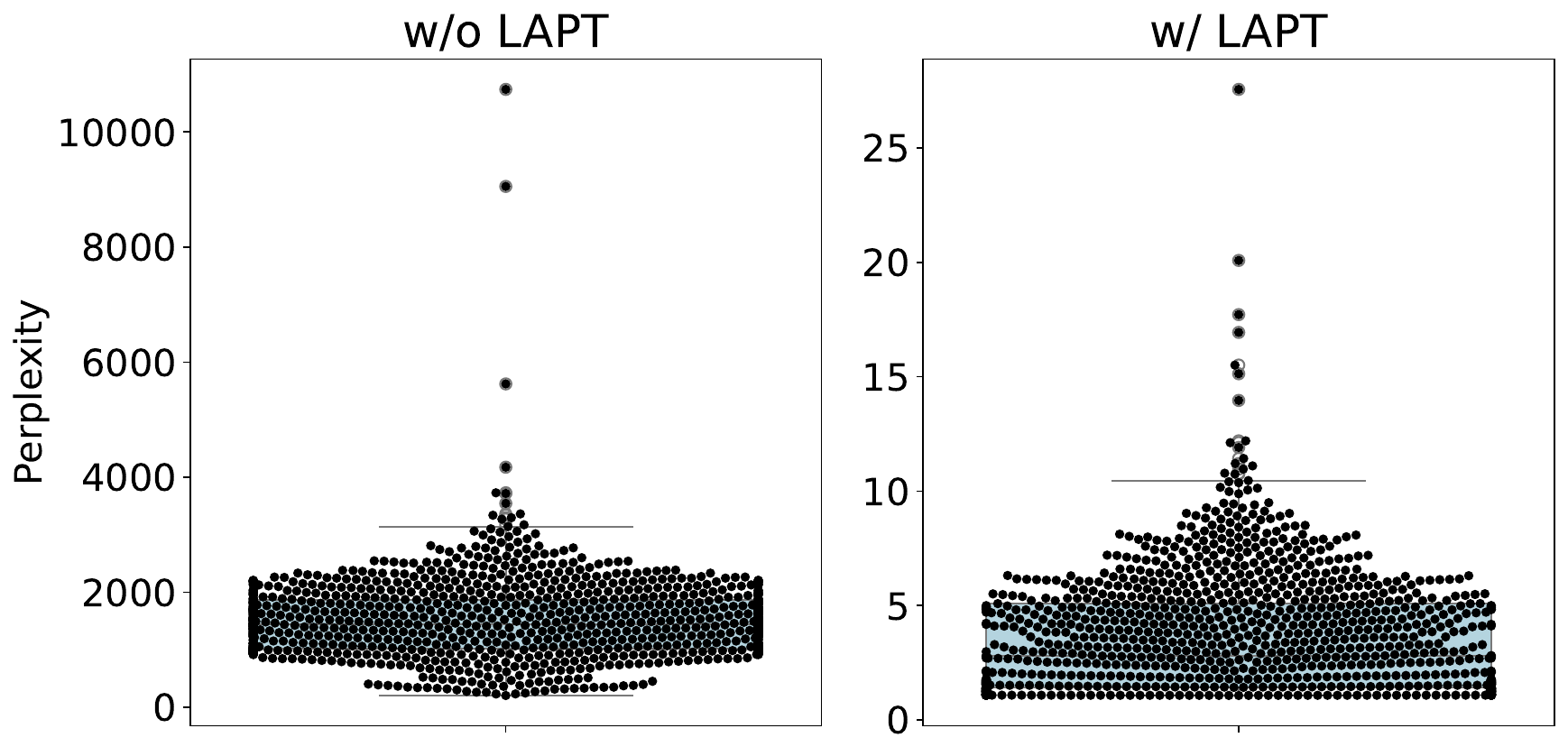}
        \caption{Sanskrit}
    \end{subfigure}
    \hfill
    \begin{subfigure}{0.3\textwidth}
        \centering
        \includegraphics[width=\textwidth]{assets/ppl_llama_khm.pdf}
        \caption{Khmer}
    \end{subfigure}
    \hfill
    \begin{subfigure}{0.3\textwidth}
        \centering
        \includegraphics[width=\textwidth]{assets/ppl_llama_mnc.pdf}
        \caption{Manchu}
    \end{subfigure}
    \hfill
    \caption{The perplexity distribution and box plot of Llama 3.1-based our proposed method.}
    \label{fig:ppl_distribution_all}
\end{figure*}

\begin{figure*}
    \centering
    \begin{subfigure}{0.3\textwidth}
        \centering
        \includegraphics[width=\textwidth]{assets/wo_removal_ppl_llama_de.pdf}
        \caption{German}
    \end{subfigure}
    \hfill
    \begin{subfigure}{0.3\textwidth}
        \centering
        \includegraphics[width=\textwidth]{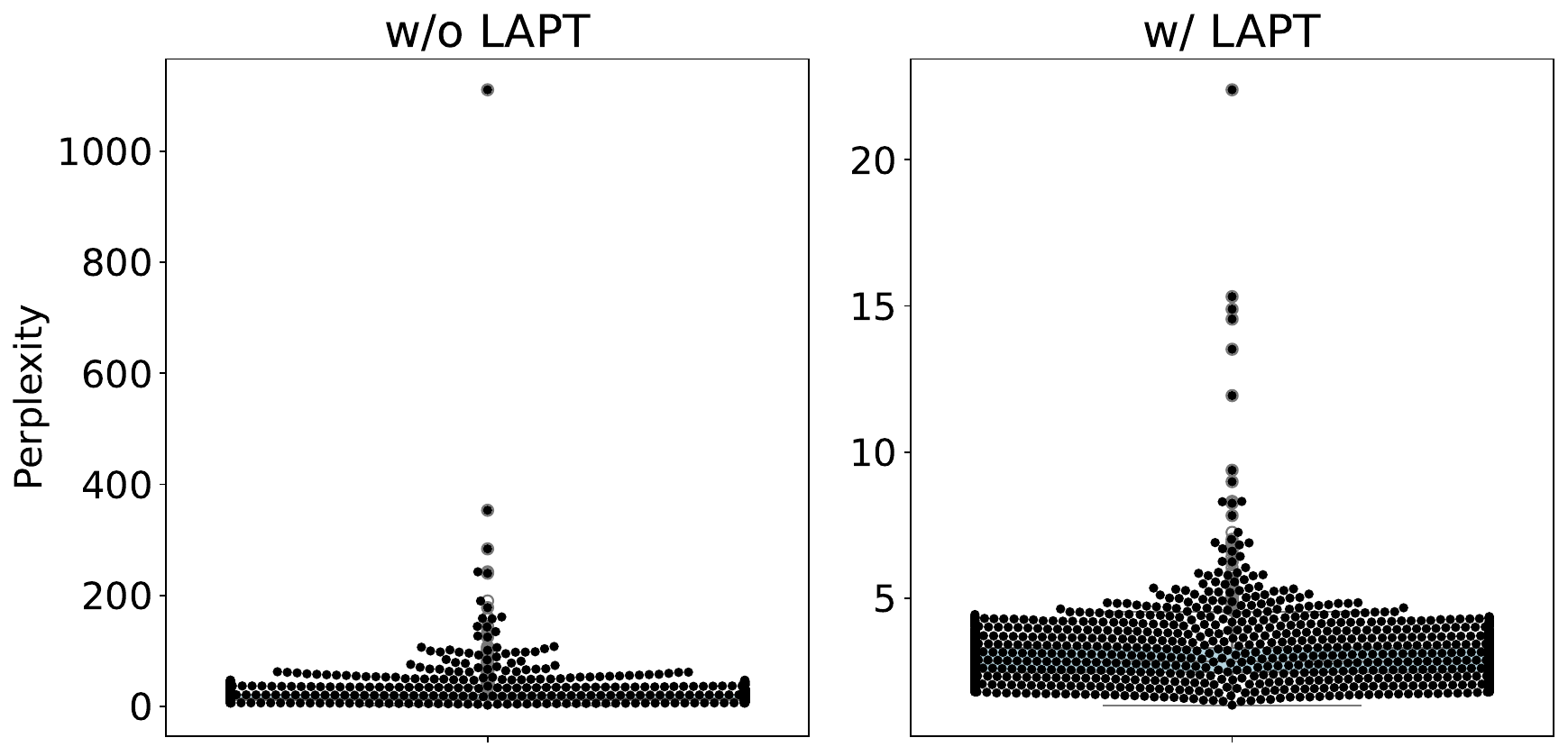}
        \caption{Japanese}
    \end{subfigure}
    \hfill
    \begin{subfigure}{0.3\textwidth}
        \centering
        \includegraphics[width=\textwidth]{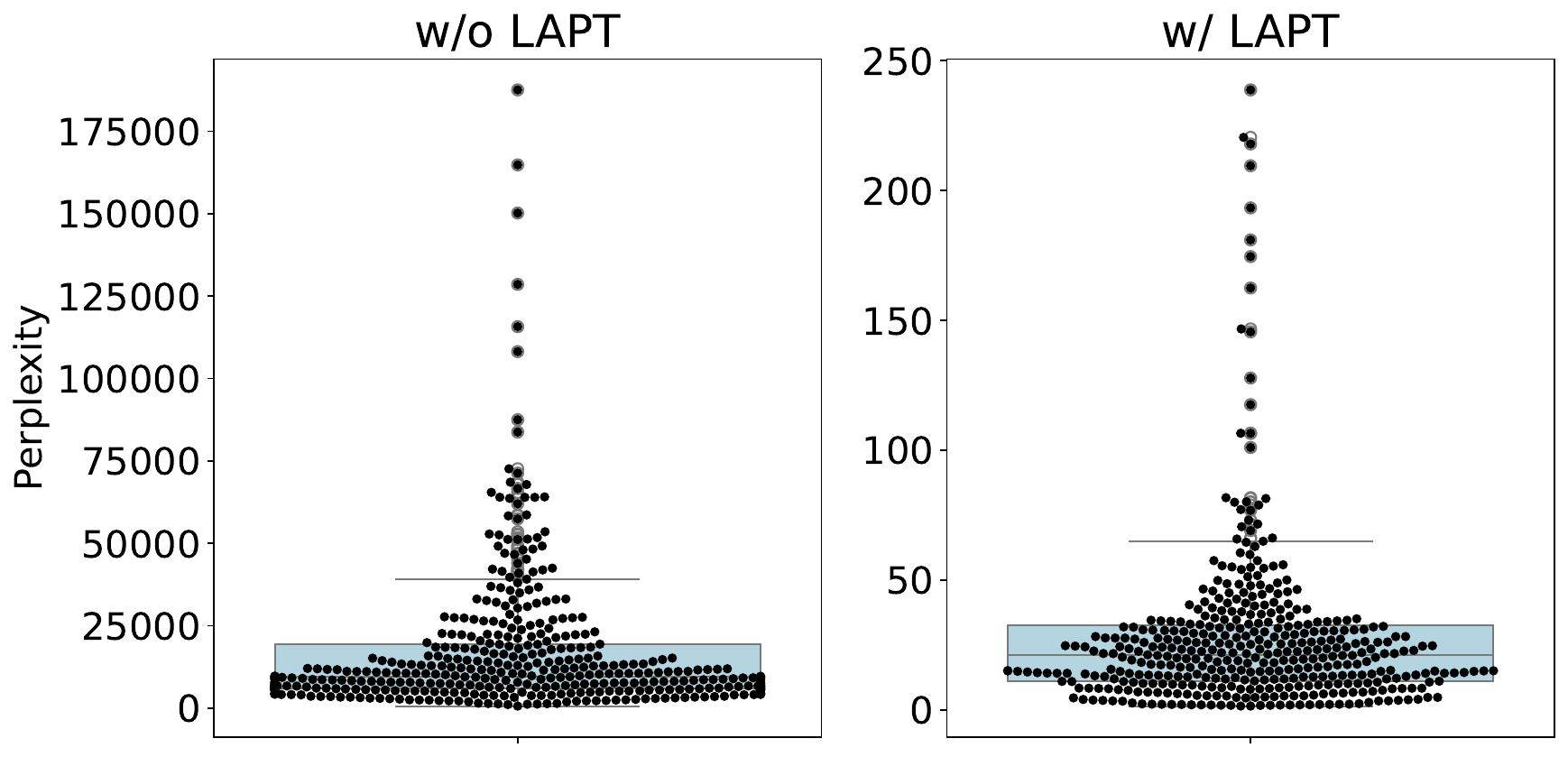}
        \caption{Old English}
    \end{subfigure}
    \hfill
    \begin{subfigure}{0.3\textwidth}
        \centering
        \includegraphics[width=\textwidth]{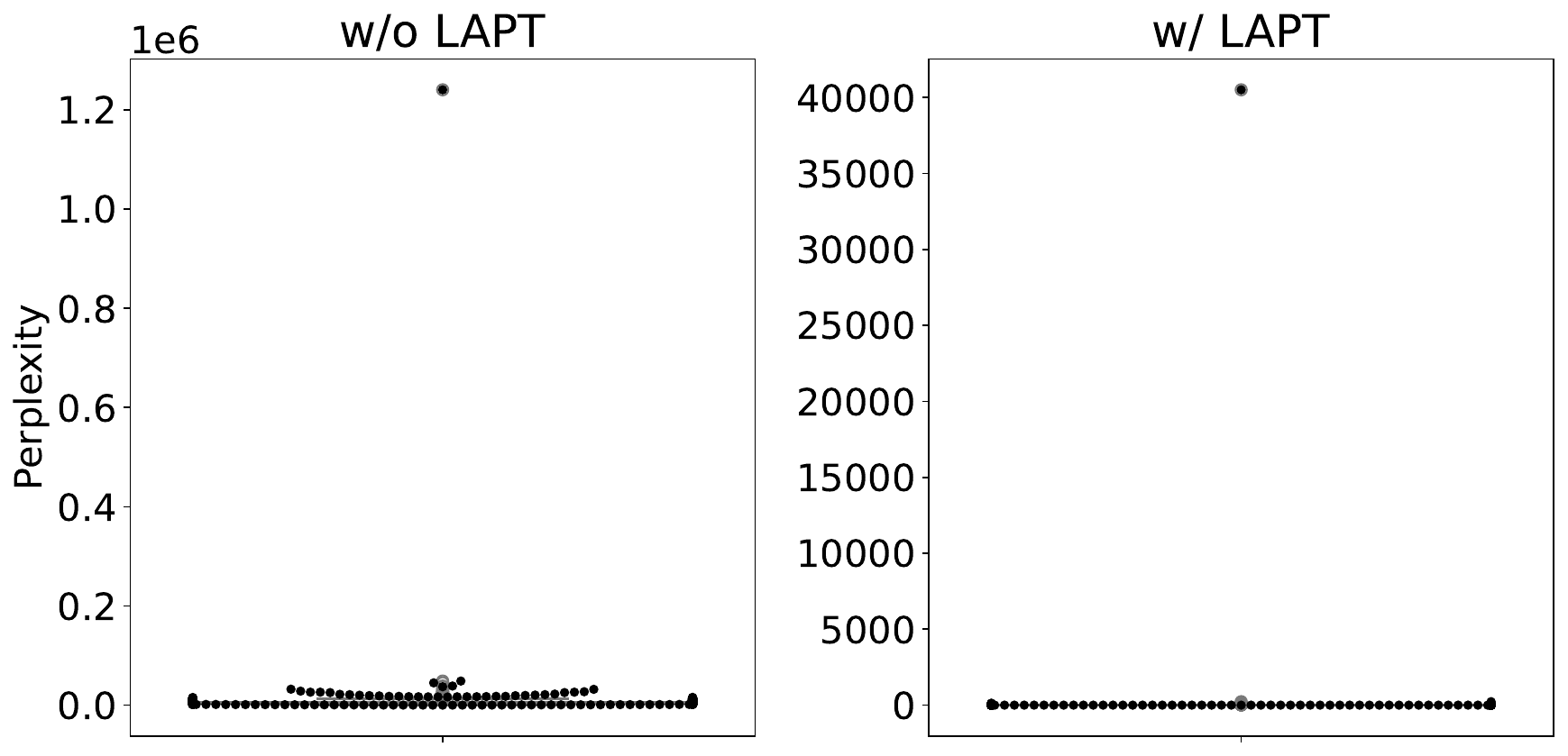}
        \caption{Uyghur}
    \end{subfigure}
    \hfill
    \begin{subfigure}{0.3\textwidth}
        \centering
        \includegraphics[width=\textwidth]{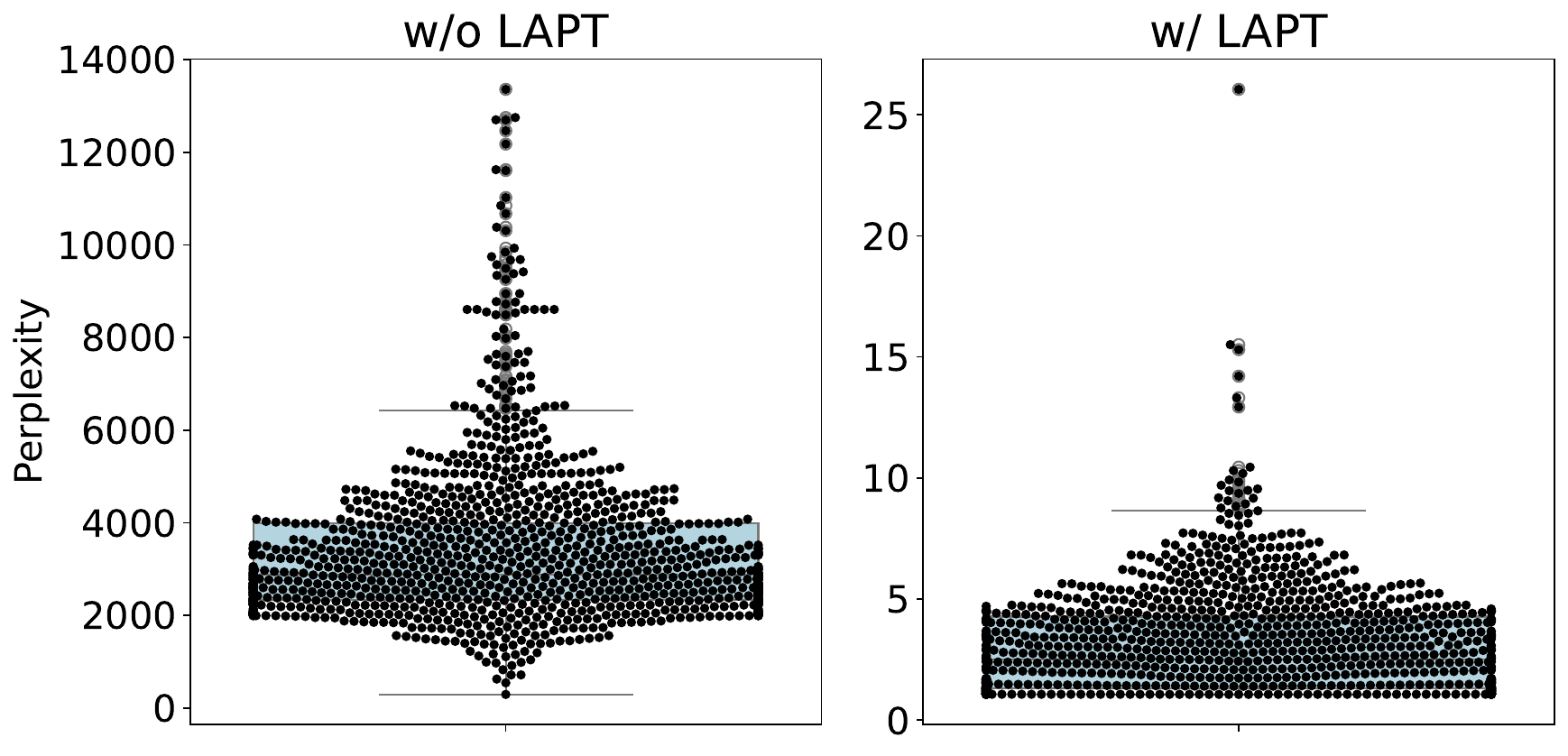}
        \caption{Sanskrit}
    \end{subfigure}
    \hfill
    \begin{subfigure}{0.3\textwidth}
        \centering
        \includegraphics[width=\textwidth]{assets/wo_removal_ppl_llama_khm.pdf}
        \caption{Khmer}
    \end{subfigure}
    \hfill
    \begin{subfigure}{0.3\textwidth}
        \centering
        \includegraphics[width=\textwidth]{assets/wo_removal_ppl_llama_mnc.pdf}
        \caption{Manchu}
    \end{subfigure}
    \hfill
    \caption{The perplexity distribution and box plot of Llama 3.1-based our proposed method without the removal.}
    \label{fig:wo_ppl_distribution_all}
\end{figure*}

Figure~\ref{fig:ppl_distribution_all} illustrates the perplexity distribution.
There are some outliers even after the LAPT was performed.
Figure~\ref{fig:wo_ppl_distribution_all} illustrates the perplexity distribution without the removal step.
The distribution is similar to that with the removal step.

\begin{table}[!t]
\centering
\small
\setlength{\tabcolsep}{2pt}
\begin{tabular}{ll}
\toprule
Language Models       &  HuggingFace ID                      \\
\midrule
RoBERTa (base)      &  FacebookAI/roberta-base         \\ 
XLM-R (base)        &  FacebookAI/xlm-roberta-base        \\ 
Llama 3.1 8B   &  meta-llama/Llama-3.1-8B       \\
Gemma 2 9B  &  google/gemma-2-9b      \\ 
\bottomrule
\end{tabular}
\caption{Lists of the models with Hugging Face IDs we used in this study.}
\label{tab:models}
\end{table}

\begin{table}[t]
    \centering
    \small
    \begin{tabular}{lr}
    \toprule
        Parameter & Value \\
         \midrule
        Batch size & 8, 16 \\
        Epochs & 2, 50 \\
        Sequence length & 256, 512 \\
        Learning rate & $1 \times 10^4$\\
        Learning rate scheduler & cosine \\
        Adam $\beta_1$ & 0.9 \\
        Adam $\beta_2$ &  0.999 \\
        Adam $\epsilon$ & $1 \times 10^8$ \\
        Precision & bf16 \\
        Seed & 42\\
    \bottomrule
    \end{tabular}
    \caption{Hyperparameters for LAPT. We train CLMs with batch size 8, 2 epochs, and sequence length 512, and MLMs with batch size 16, 50 epochs, and sequence length 256.}
    \label{tab:params_lapt}
    \end{table}

\begin{table}[!t]
    \centering
    \small
    \begin{tabular}{lr}
    \toprule
        Parameter & Value \\
    \midrule
        Batch size & 64\\
        Epoch & 25 \\
        Learning rate & 5e-5\\
        Learning rate scheduler & linear \\
        Adam $\beta_1$ & 0.9 \\
        Adam $\beta_2$ &  0.999 \\
        Adam $\epsilon$ & $1 \times 10^8$ \\
        Seed & 42 \\
    \bottomrule
    \end{tabular}
    \caption{Hyperparameters for task adaptation.}
    \label{tab:params_finetuning}
\end{table}

\end{document}